\pdfoutput=1

\documentclass[11pt]{article}

\usepackage[final]{acl}

\usepackage{times}
\usepackage{latexsym}

\usepackage[T1]{fontenc}

\usepackage[utf8]{inputenc}

\usepackage{microtype}

\usepackage{inconsolata}

\usepackage{graphicx}

\title{Investigating the Robustness of Retrieval-Augmented Generation at the Query Level}

\author{
  \textbf{Sezen Perçin\textsuperscript{1}\thanks{Work performed while at Intel Labs.}},
  \textbf{Xin Su\textsuperscript{2}},
  \textbf{Qutub Sha Syed\textsuperscript{2}},
  \textbf{Phillip Howard\textsuperscript{3}},
  \textbf{Aleksei Kuvshinov\textsuperscript{1}},\\
  \textbf{Leo Schwinn\textsuperscript{1}},
  \textbf{Kay-Ulrich Scholl \textsuperscript{2}}
\\
  \textsuperscript{1}Technical University of Munich,
  \textsuperscript{2}Intel Labs,
  \textsuperscript{3}Thoughtworks,
\\
  \small{
    \{sezen.percin, aleksei.kuvshinov, l.schwinn\}@tum.de, \{xin.su, syed.qutub\}@intel.com, phillip.howard@thoughtworks.com
  }
}

\begin{document}
\maketitle
\begin{abstract}
Large language models (LLMs) are very costly and inefficient to update with new information. To address this limitation, retrieval-augmented generation (RAG) has been proposed as a solution that dynamically incorporates external knowledge during inference, improving factual consistency and reducing hallucinations. Despite its promise, RAG systems face practical challenges-most notably, a strong dependence on the quality of the input query for accurate retrieval. In this paper, we investigate the sensitivity of different components in the RAG pipeline to various types of query perturbations. Our analysis reveals that the performance of commonly used retrievers can degrade significantly even under minor query variations. We study each module in isolation as well as their combined effect in an end-to-end question answering setting, using both general-domain and domain-specific datasets. Additionally, we propose an evaluation framework to systematically assess the query-level robustness of RAG pipelines and offer actionable recommendations for practitioners based on the results of more than 1092 experiments we performed.
\end{abstract}

\section{Introduction}
\label{sec:intro}

Recent advancements in the capabilities of large language models (LLMs) have revolutionized the field of natural language processing (NLP) and have achieved impressive performance across a broad range of downstream applications. Their success can largely be attributed to the massive text datasets on which they are trained and their increasing size in terms of model parameters. However, these factors that have enabled their success also limit their practical implementation in downstream applications. For example, a business seeking to implement an LLM to answer questions about proprietary internal documents may lack the compute resources and dataset scale needed to train an LLM with the necessary domain knowledge.

Even when an LLM can be properly trained on domain-specific data, all existing models are prone to the well-known issue of hallucination \citep{domspecifichallucination}. This phenomenon, where LLMs produce confident-sounding, factually inaccurate responses, is particularly problematic for applications in which downstream users may lack the necessary domain knowledge to identify and correct the inaccuracies. Further compounding this issue is the inherent lack of transparency in how LLMs arrive at their generated responses \citep{llmtransp}.

Retrieval-augmented generation (RAG) has been proposed as a solution for mitigating the aforementioned shortcomings of LLMs \citep{lewis2020retrieval,ram2023context}. Specifically, a RAG system utilizes a text retriever to identify the documents in a text corpus most relevant to a given query via a semantic similarity measure. The most similar retrieved documents are then provided as additional context to the LLM, along with the query for context-augmented generation. By conditioning generation on retrieved documents, new information can be incorporated into LLMs’ responses without additional training. Furthermore, RAG reduces the likelihood of hallucinations by grounding generation in documents which are a trusted source of truth and enables greater transparency by allowing end-users to inspect documents which were used to produce the response generated by an LLM.

\begin{figure*}[h]
\centering
\centerline{\includegraphics[width=1\textwidth]{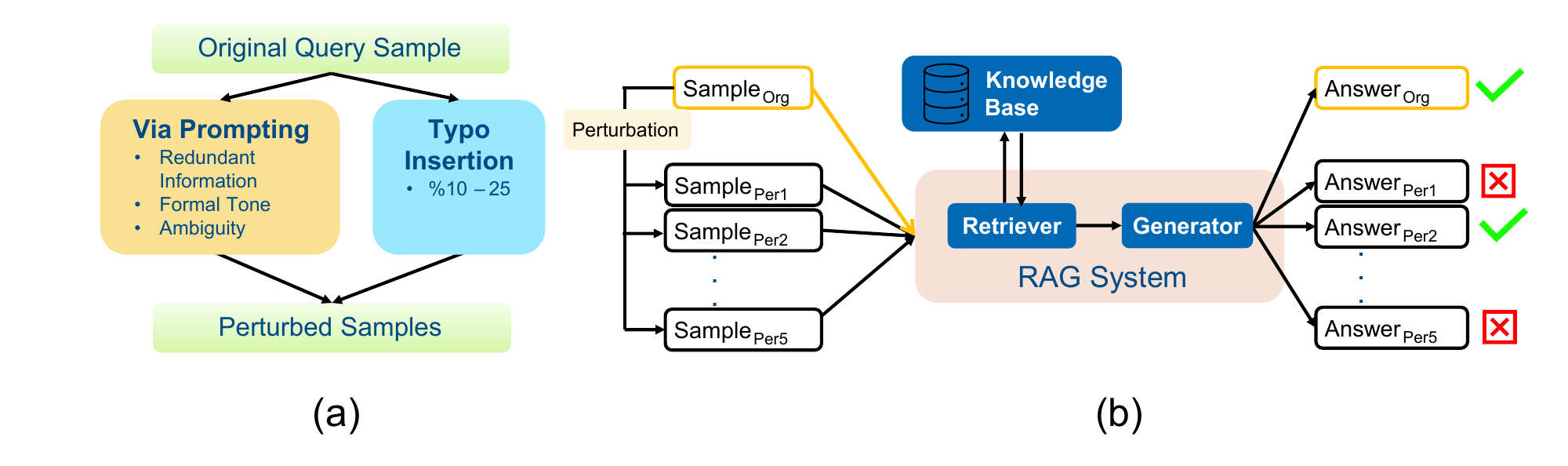}}
\caption{Illustration of our approach to evaluating RAG robustness. (a) Perturbations are generated via prompting an LLM  or random insertion of typographical errors. (b) Evaluation datasets are formed using five perturbed samples for each original example. (Org: Original, Per: Perturbed)}
\label{perturbed_versions}
\end{figure*}

While RAG systems have achieved impressive
performance, an essential question for their practical application in downstream systems is how 
variations in a user’s query impact the relevancy of 
retrieved results. For instance, different users seeking the same information may phrase their queries
differently or introduce typographical errors to the
query. A desirable attribute of a RAG system is
that the elements in the system (e.g., retrievers) are
robust to such variations and perform similarly for
all users. This is important for overall usability and fairness, as how humans phrase a question can reflect differences in educational and cultural backgrounds.

In this work, we systematically investigate the sensitivity of RAG systems to perturbations in their input queries. Specifically, we introduce transformations and varying levels of typographical errors to queries across several benchmark datasets, measuring how such perturbations impact the performance of different components in a RAG system. Across 4 popular retrievers, we find consistent variations in their performance in the face of our query perturbations.

Moreover, we investigate the correlations between the performances of each module and joint pipeline and provide insights on the decoupling of the case-specific sensitivities arising from each module. Motivated by these findings, we provide recommendations for improving RAG system robustness to query variations and propose an evaluation framework. To our knowledge, this is the first work providing a framework to decouple each module’s sensitivities in RAG pipelines for robustness research.

To summarize, our contributions are as follows:
\begin{enumerate}
    \item We introduce a framework for measuring the robustness of RAG systems to varying levels of typographical errors and frequently occurring prompt perturbation scenarios for input queries.
    \item We conduct experiments using 4 different retrievers and 3 different LLMs, evaluating 12 resulting question-answering pipelines in total. Further, we cover datasets of different characteristics and domains to provide a comprehensive analysis.
    \item Based on our experimental results and additional analyses, we provide insights and recommendations for improving the robustness of RAG systems.
\end{enumerate}

We will make our data and code publicly available to support future work on evaluating the robustness of RAG systems to variations in user queries.

\section{Related Work}
Existing studies on RAG robustness can be broadly grouped as focusing on the retriever, on the LLM as the final generator, or on the entire RAG pipeline.

\paragraph{Retriever-Level Robustness} 
Research in this category explores how retrievers maintain performance under various query perturbations \citep{neuralIR_robustness, charbert, sidiropoulos2022analysing, oodqueryrobustness1, liu2023robustness, arabzadeh2023noisy}. For example, \citet{charbert} explores how BERT-based retrievers cope with spelling errors and proposes encoding strategies and training procedures to improve robustness. In contrast, \citet{liu2023robustness} studies how generative retrievers handle query variants. Meanwhile, \citet{arabzadeh2023noisy} evaluates retriever stability by perturbing queries’ dense representations directly.

\paragraph{LLM-Level Robustness} 
Another body of work targets the LLM itself, examining how effectively the model filters out irrelevant or misleading context \citep{llm_robustness1} and how it responds to perturbations in the prompt at various granularity levels, from character-level to entire sentences \citep{llm_robustness2,promptbench}. These studies primarily aim to ensure that the model’s outputs remain accurate and consistent despite possible noise or adversarial modifications in the prompt.

\paragraph{Pipeline-Level Robustness}
A further line of research adopts a holistic view of RAG, focusing on how noise in retrieved documents-such as irrelevant passages or misinformation-affects overall performance, proposing methods to mitigate these issues \citep{rag_robustness1, rag_robustness5, rag_robustness2, rag_robustness3, rag_robustness4, shen2024assessing, han2023robustqa}. For example, \citet{rag_robustness1} tests whether the model can ignore non-relevant content or misinformation and, if necessary, refuse to answer when the retrieved context is unreliable. Approaches such as \citet{rag_robustness5} and \citet{rag_robustness3} investigate various types of erroneous or irrelevant information in RAG and introduce new training techniques to counteract performance degradation. In addition, \citet{rag_robustness4} considers scenarios in which some retrieved documents may have been maliciously altered, presenting a defense mechanism.

Despite these efforts, many studies focus on either the retriever or the overall RAG workflow without systematically analyzing how each component behaves under diverse query perturbations. By contrast, we conduct a more comprehensive analysis spanning the entire RAG pipeline and propose a new framework that offers a clearer, more intuitive assessment of system robustness.

\begin{table}
  \centering
  \begin{tabular}{lcc}
    \hline
    \textbf{Dataset} & \textbf{PERT}& \textbf{Corpus}\\
    \hline
    NQ  & 1496 &2.68M \\
    HotpotQA  & 1494&5.23M\\
    BioASQ  & 378 &14.91M\\\hline
  \end{tabular}
  \caption{Number of samples and the size of the corpora for each dataset used in the experiments: HotpotQA \citep{dataset_hotpotqa}, NQ\citep{dataset_nq} and BioASQ\citep{dataset_bioasq}.(PERT: Number of perturbed samples for each perturbation type)}
  \label{sample-table}
\end{table}

\section{Data Perturbations}

We investigate strategies to systematically evaluate the robustness of the RAG pipeline under different input perturbations that commonly appear in real-world applications. For each type of perturbation, we also quantify how the performance of different modules in the RAG pipeline changes. We focus on perturbations that do not significantly alter the semantic meaning of the query in practical RAG use cases while having a high chance of occurrence. Specifically, given an original query \(q\), we apply a perturbation \(\mathrm{Perturb}(q)\) such that \(\mathrm{Perturb}(q)\) retains the same or very similar semantics as \(q\). In this work, we generate the perturbed samples in two ways: either via prompting an LLM or by inserting random typos. 

\subsection{Perturbations Via Prompting}
To enable large-scale evaluation of the perturbations, this first category uses the LLM GPT-4o as a data generator to produce synthetically perturbed samples. This approach is motivated by the observation that LLMs are very successful at processing textual input and are widely used for the generation of synthetic data as well as adversarial examples. Additional details on the evaluation of the generated samples, along with the prompts used to generate them, can be found in Appendix \ref{app_aut_sample_gen}. 

We investigate three under-explored query perturbations in the context of RAG. For a query taken from the HotpotQA dataset, we provide examples corresponding to each perturbation. The original non-perturbed sample is shown below.

\begin{center}
\textit{"when does the cannes film festival take place"}
\end{center}

\paragraph{Redundancy Insertion} This perturbation type reflects the cases where a user inserts elements into their queries which do not add additional value or information that will help the system in response generation.

\textit{"I’m curious to know the specific dates or time frame for the Cannes Film Festival, an internationally renowned event that celebrates cinema and attracts filmmakers, actors, and industry professionals from all over the globe to the picturesque city of Cannes in France."}

\paragraph{Formal Tone Change} This perturbation refers to the scenarios where the input queries are expressed in a more formal manner than the general case. This transformation does not lead to semantic meaning changes in the overall query while leading to variances on the surface level.

\begin{center}
\textit{"When is the Cannes Film Festival set to be held?"}
\end{center}

\paragraph{Ambiguity introduction} This perturbation covers the possibility of expressing the input queries in a way that is unclear or open to interpretation in many ways. This can be done by, for example, inserting words such as "might" into the formulations of the sentences or changing words to more general correspondents (e.g., changing "rapper" to "artist").

\begin{center}
\textit{"When might the Cannes Film Festival be held?"}                                                                                             
\end{center}

\subsection{Typo Perturbations}
It is also common for users to make minor spelling mistakes, especially when typing quickly. In many cases, these errors do not impede human comprehension-for example, typing ``tomrow'' instead of ``tomorrow''. Nevertheless, such typographical errors may still affect retrieval and generation in a token-based RAG pipeline. One potential solution is to run a dedicated spell checker before feeding the text into the RAG pipeline, but this introduces additional computational overhead and may be inaccurate for domain-specific terminology. For instance, the term ``agentic,'' recently popularized in AI discussions, often triggers false alarms in existing spell-check systems.

To explore the effect of spelling errors, we use the TextAttack~\citep{textattack} library to simulate minor typos by replacing characters in the query based on their proximity on a QWERTY keyboard. We experiment with perturbing 10\% and 25\% of the words in each query to ensure the overall intent remains understandable. In addition, we maintain a stop-word list that remains unaltered to preserve key semantic content. Example obtained with typo perturbations at 10\% and 25\% levels in respective order are provided below.

\begin{center}
\textit{"when does the cannes film festival take plac"}
\end{center}
\begin{center}
\textit{"when does the cannes fjlm festival takr place"}
\end{center}

For each perturbation type, we take each sample from the original dataset and generate 5 new perturbed samples based on the original sample. We present an overview of our approach in Figure \ref{perturbed_versions}.

\section{Experiment Details}
In this section, we describe the elements used in these experiments, such as the datasets and models, to assess the robustness of the RAG systems. 

\subsection{Datasets}

We use the widely adopted retrieval benchmark BEIR~\citep{BEIR}. Since not all of the tasks are suitable for the RAG setting, we focused on the task of question-answering. Out of three datasets in the "Question-Answering" (QA) category of the benchmark, we chose NQ and HotpotQA since these datasets have short answer labels in the form of a few keywords. This decision eases the evaluation process while enabling for a more stable robustness analysis. Moreover, we include BioASQ from the "Bio-Medical IR" category to see the effect of the perturbations on a domain-specific QA dataset. Similar to \citet{ragged}, we integrated datasets having different corpora (Wikipedia and biomedical), characteristics (multi-hop, single-hop) and sizes.

\subsection{Models}
In order to assess the robustness of the RAG pipeline against query perturbations, we define our RAG pipeline to consist of a retriever and a generator, as shown in Figure \ref{perturbed_versions}. In this system, the retriever is responsible of interacting with a knowledge base to retrieve most relevant documents conditioned on the given query, while the generator produces a final response using the initial query along with the retrieved context information.

\begin{figure*}[h]
\centering
\centerline{\includegraphics[width=1\textwidth, height=0.3\textwidth]{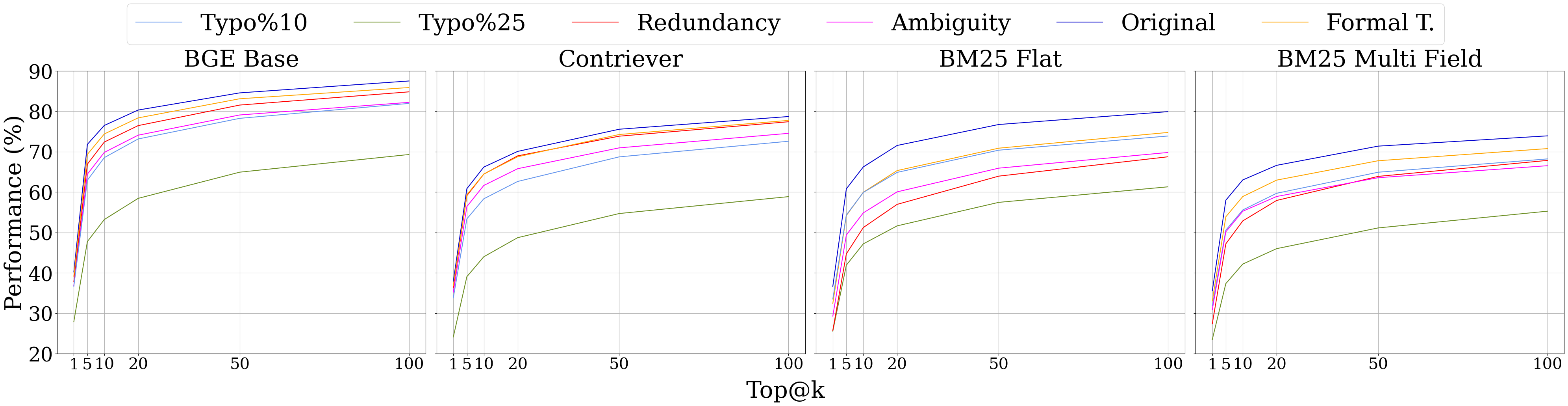}}
\caption{Recall@k results obtained with different retrievers on HotpotQA with respect to the changing "k" parameter as shown in axis Top@k.}
\label{prec_vs_recall}
\end{figure*}

\paragraph{Retriever}

We employ three main retrievers in our system: BGE-base-en-v1.5~\citep{bgebase}, Contriever~\citep{contriever} as dense retrievers, and BM25~\citep{BM25} as a sparse retriever. For BM25, we adopt two variants: one that considers only the document content and another that uses a multi-field setup including both the document content and the ``Title'' field. We obtain the publicly available precomputed indices from the Pyserini framework~\citep{pyserini}.

\paragraph{LLM Generator}
As generators, we used widely employed LLMs between 7-8B parameters in size: Llama-3.1-8B-Instruct \citep{llama31}, Mistral-7B-Instruct-v0.2 \citep{mistral7b} and Qwen2.5-7B-Instruct \citep{qwen2.5}. Using the BERGEN framework \citep{bergen}, we set the maximum input length to 4096 tokens, the maximum generated tokens to 128, and the temperature to 0. When generating the responses, we used greedy decoding. Following the setup provided by the framework, when incorporating the retrieved documents into the LLM’s input, we truncate each document to a maximum of 100 words. We use vLLM~\citep{vllm} as our inference framework to run these models. All the experiments are performed on a NVIDIA GeForce RTX 3090 GPU.

\subsection{Standard Evaluation Metrics}

To evaluate the performance of retrievers, we utilized a widely employed metric for assessing the information retrieval of dense and sparse retrievers, namely the  \textit{Recall@k}, where the parameter k defines the top ``k'' documents that are returned by the retriever. While investigating retriever robustness, we experimented with different $k$ choices; however, during the end-to-end experiments we define $k$ as 5. To evaluate the LLM-generated content in the RAG pipeline, we adopted a surface matching metric from the BERGEN framework~\citep{bergen}, called \textit{Match}. This metric checks whether the generated output contains the answer span. 

Unlike recent trends that use an LLM for automated evaluation, we opt for a model-free assessment to ensure robust and reproducible analysis and to avoid fluctuations caused by changes in the evaluating LLM. Moreover, it is intended that our evaluation framework avoid the computational cost associated with employing an LLM-based evaluator, thereby removing the need to choose a model that is parameter-efficient while ensuring evaluation quality.

\section{Experiments}

In this section, we detail the steps in our analysis framework and describe our findings in the designed experiments. 

\subsection{Our Analysis Framework}

To understand the effect of each query perturbation on the RAG pipeline, we first perform isolated assessments on each module. For retrievers, we examine the changes in performances measured in Recall@k on the text passage retrieval task. For generators, we define two settings to cover two mechanisms that an LLM can rely on to generate answers. Then we move to the end-to-end pipeline and analyze the effect of each perturbation on the overall RAG performance. We further provide analysis on correlations to individual module sensitivities and changes in internal LLM representations. Details of each experiment are provided in the following sections.

\subsection{Retriever Robustness}
The analysis of the RAG pipeline begins with the retriever component, which interacts with a knowledge base to return a list of ranked elements conditioned on the input query. This knowledge base, consisting of text passages, will be referred to as "documents" in this study.

To investigate the robustness of the retrievers, we analyzed the performance changes observed with each perturbation. The resulting effects of perturbation types using different retrievers on the HotpotQA dataset are shown in Figure ~\ref{prec_vs_recall}. We also provide results for the remaining retriever and dataset combinations in Figure \ref{fig: remaining_retriever_results}.

Our analysis and the recall curves show that the dense retrievers are more robust against the redundant information contained when compared to the sparse methods, however sparse methods performances are more robust against the typos introduced to the input queries. Formal tone change is the least effective perturbation types on the retriever performances across both retrieval categories. While increased typo levels i.e. \%25 lead to least performance scores across all combinations in general, redundant information insertion leads to even worse performances for sparse retrievers when used for the domain specific BioASQ dataset.

\subsection{Generator Robustness}

In order to assess how different generators handle different types of query perturbations, we examine the performance changes caused by each perturbation type. These performance changes are investigated in two settings representing the two abilities the LLMs can use in QA task to generate an answer. First, they can use their parametric knowledge gained during pretraining to answer the input queries. Second, these models use their context utilization abilities to integrate the knowledge given in their context window into the generated answer. We refer to these settings as "closed-book" and "oracle" (respectively).

\begin{figure}[h]
\centering
\centerline{\includegraphics[width=0.5\textwidth]{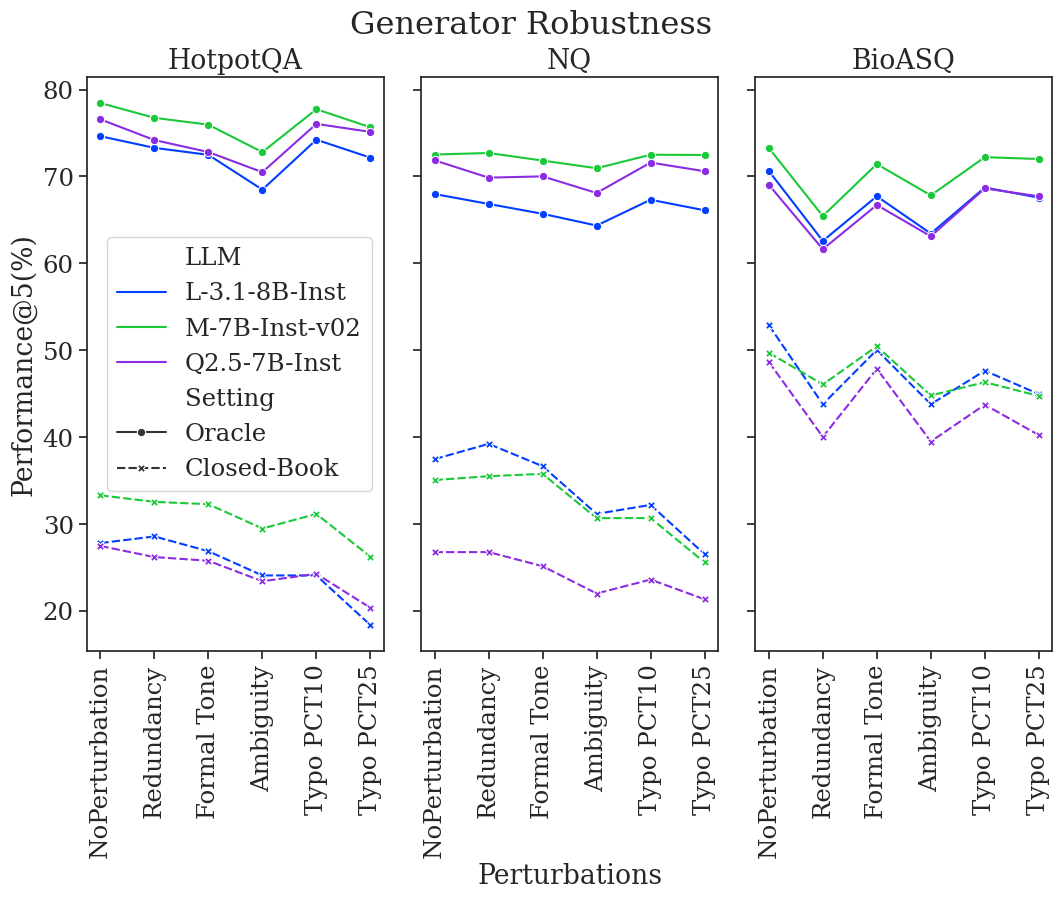}}
\caption{LLM performances under different perturbations using the "Match" metric in closed-book and oracle settings.}
\label{gen_robustness}
\end{figure}

Closed-book experiments require the generator LLM to answer the given queries without accessing any external knowledge source and hence completely relying on the knowledge stored in their parametric memory. In contrast, in oracle experiments, the existence of an "oracle" retrieval system is assumed to return only correct documents and nothing else. This setting establishes an upper bound for the system, as the models have access to only correct information and no other information. For each dataset, we report the experimental results in Figure \ref{gen_robustness}, where each generator is differentiated by color and different settings are reflected by line styles. All results are reported in Match metric for the original and perturbed datasets.

Our results show that the generator robustness is dependent on the nature of the dataset and the sensitivity of each LLM against difference perturbations. The LLMs tend to follow similar trends on a dataset and the perturbations result in performance drops in general. However, there are cases where LLMs are behaving differently. For example, while all perturbations result in performance reductions, the redundant information increases the performance of Llama 3.1-8B-Instruct in certain cases when parametric knowledge is incorporated.  Similarly, the formal tone change causes performance decreases and increases based on the LLM and the dataset chosen.

When perturbation types are individually assessed, ambiguity insertion decreases model performance in both settings across different datasets, posing a challenge for LLMs. While redundant information has a low impact on performance for general datasets such as NQ and HotpotQA, it causes drastic performance drops on the domain-specific BioASQ dataset in both settings.

Moreover, the typo insertions are particularly impactful in the closed-book setting, resulting in great performance decreases. In contrast, when the necessary knowledge is provided, the systems are mostly able to recover from these perturbations, especially at a level of 10\%. This indicates that when combined with information, the query perturbations result in different impacts than the effects seen in closed book settings which are commonly used to evaluate LLM robustness.

\begin{figure*}[h]
\centering
\centerline{\includegraphics[width=1\textwidth]{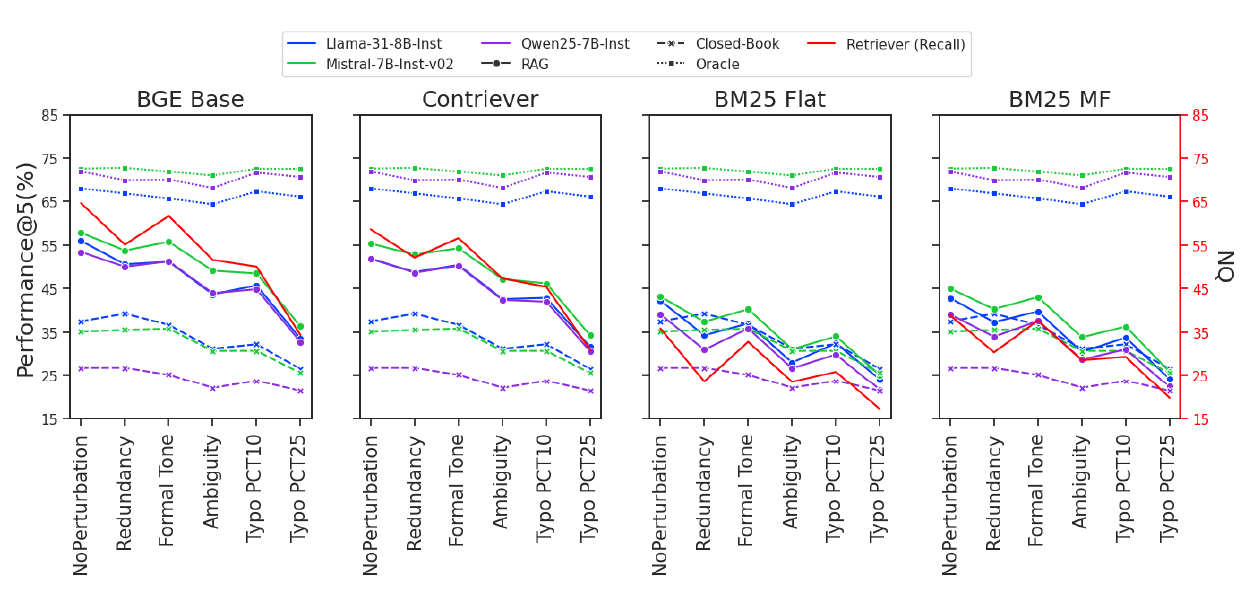}}
\caption{The average end-to-end results on NQ dataset according to "Match" metric.}
\label{rag_results_hotpotqa}
\end{figure*}

Lastly, the performance of the models in the closed book setting is not reflected in the oracle performances, which underlines the importance of the context utilization abilities and the retrieval incorporated to the RAG pipelines. For instance, although the parametric knowledge of Mistral-7B-Instruct-v0.2 varies across datasets, it is the best performing model in the oracle setting. 

\subsection{RAG Robustness}
Finally, we analyze the joint effect of combining different elements to form an end-to-end RAG system. This setting differs from the oracle experiments defined earlier in that the system includes a non-ideal retriever which can return irrelevant documents. 

Figure \ref{rag_results_hotpotqa} and \ref{all_rag_match} display the average end-to-end results of the pipeline reported in "Match" metric. Each window incorporates a single retriever's data while different generator combinations are colored accordingly. The red curve on the plots shows the retrieval scores using the Recall@5 metric while the horizontal axis shows the perturbation types.

These results indicate that the performance trends observed under various perturbations are predominantly characterized by the performance of the retriever. This is most evidently demonstrated in the case of the NQ dataset, as illustrated in Figure \ref{rag_results_hotpotqa}, where the RAG outcomes manifest as retriever performance trends. However, it is evident that the retriever performance is not fully reflected in the end-to-end performance on the BioASQ dataset when BGE Base or Contriever is used in combination with different LLMs, as shown in Figure \ref{all_rag_match}. In this scenario, despite the low performance changes observed in the retriever performance, RAG performance exhibits significant declines, particularly in the cases of ambiguity and redundancy introductions. To further explore the underlying causes of these observations, we conduct a more in-depth analysis. 
\paragraph{Correlation to the Individual Module Performances:} 
We investigated the Pearson correlation scores between the retriever, generator, and end-to-end performances. First, the performance discrepancy between each perturbed sample and its original non-perturbed counterpart was determined using the metrics "Recall@5" for retrievers and "Match" for generator and end-to-end performances. The Pearson correlation coefficients calculated for retrieval-RAG and generator-RAG settings can be found in Table \ref{pearsoncorr} for the BGE-Base retriever in combination with Llama 3.1-8B-Instruct for the BioASQ and NQ datasets. The results obtained with different modules are reported in Table \ref{pearsoncorr_all_bioasq} and \ref{pearsoncorr_all_nq}.

The correlation scores indicate that different dominant factors exist within the pipeline for different perturbation types. For example, in the case of BioASQ dataset for instances involving typo perturbations, the end-to-end results demonstrate a stronger correlation with retriever performance. Conversely, in cases of ambiguity, formal tone change, and redundancy insertion, generator-only settings exhibit higher scores. When these findings are compared to the coefficients calculated on the NQ using the same pipeline, we see that the results on NQ correlate more with the retriever performance differences. This also validates our observations that the results for the NQ dataset are mainly defined by the retriever trends. The findings of this study demonstrate the potential of such an analysis to assist practitioners in identifying the module within their pipeline that exhibits particular sensitivity to a specific perturbation types. 

\paragraph{Internal LLM Representations:} 
Lastly, we inspected the internal representations of the LLMs and analyzed how they differ when faced with various perturbations. For this analysis, we focused on the BioASQ dataset in oracle and RAG settings with BGE-Base as the retriever. We gathered the inputs given to the LLM and obtained an internal representation for these inputs by averaging over all attention heads of Llama-3.1-8B-Instruct for the last hidden state layer calculated for the last non-padding token. As the vLLM framework utilized in the experimental setup does not permit straightforward access to the internal representations of the LLMs, we obtained them by employing the Huggingface deployment of these models and evaluated the results again using BERGEN~\citep{bergen} framework.

\begin{table}[]
\begin{tabular}{|c|c|c|c|c|c|}
\hline
Type     & R & F& A& T10 & T25\\
\hline
\multicolumn{6}{|c|}{BioASQ}\\
\hline
RET   &0.05	&0.04	&0.15	&0.21$\uparrow$	&0.23$\uparrow$\\
\hline
CB   & 0.21	&0.08	&0.23	&0.05	&0.10\\
\hline
OR      &0.35$\uparrow$	&0.15$\uparrow$	&0.33$\uparrow$	&0.04	&0.12 \\
\hline
\multicolumn{6}{|c|}{NQ}\\
\hline
RET   & 0.31$\uparrow$&0.27$\uparrow$&0.30$\uparrow$&0.35$\uparrow$&0.40$\uparrow$\\
\hline
CB & 0.03&0.04&0.11&	0.08&0.16\\
\hline
OR       & 0.11&	0.14&0.15&0.06&0.03 \\
\hline
\end{tabular}
\caption{Pearson correlation coefficients for BioASQ and NQ dataset and BGE Base as retriever. (R: Redundancy, F: Formal Tone, A: Ambiguity, TX: Typo \%X; Correlations:  RET: Retriever-RAG, CB: Closed Book-RAG, OR: Oracle-RAG)}
\label{pearsoncorr}
\end{table}

\begin{figure*}[t]
  \includegraphics[width=1\linewidth]{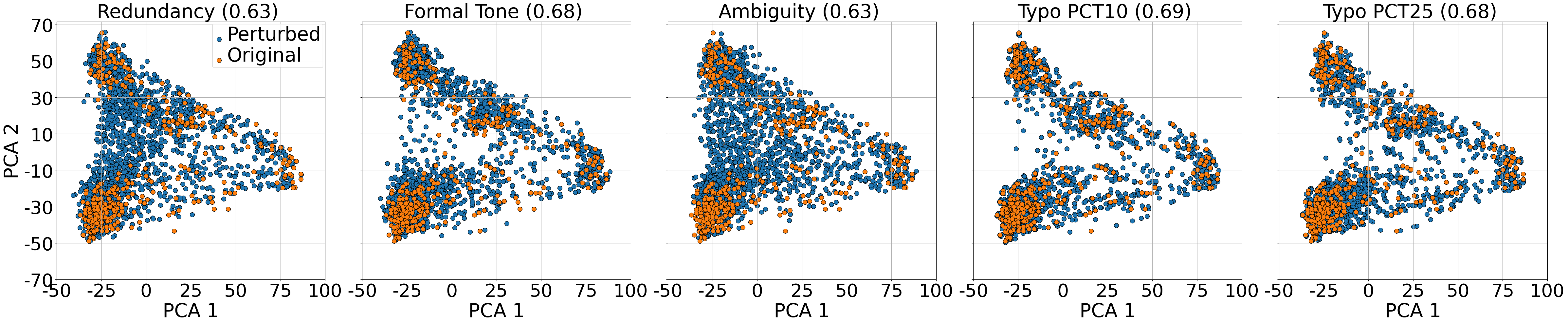} \hfill
  \includegraphics[width=1\linewidth]{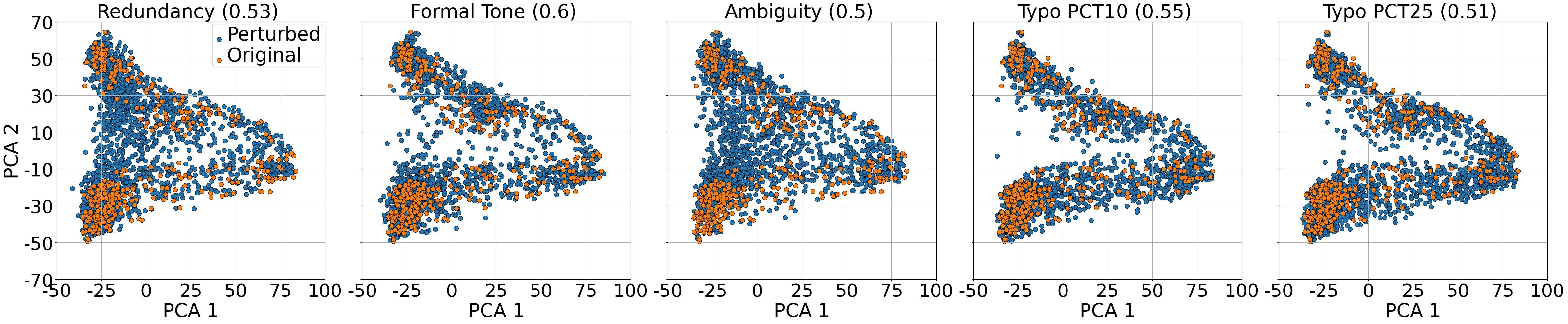}
  \caption{Representation of samples taken from Llama-3.1-8B-Instruct for the BioASQ dataset for the oracle (upper) and RAG with BGE Base (lower) settings. The Match performance of the original non-perturbed performances for oracle and RAG (with BGE Base) are 0.71 and 0.61 respectively.}
  \label{llm_emb}
\end{figure*}

We visualize these representations by projecting them onto a two-dimensional space using PCA for dimensionality reduction. The representations are shown in Figure \ref{llm_emb} for different types of perturbations. For both settings, we observe similar trends where the introduction of redundancy and ambiguity results in more scattered internal representations with respect to the original dataset. Only the queries vary in the oracle setting, as the documents inserted are identical across all runs. These results show that the perturbations in queries scatter the internal representations despite the existence of golden documents.

\section{Recommendations}

As a results of our broad experiments across different retrievers, generator models (i.e. LLMs) and data perturbation types, we provide evidence based practical recommendations to for the improvement of retrieval augmented generation pipelines. These insights are designed to help developers assess the robustness of their RAG pipelines against different input transformations, essentially helping developers make robustness-aware decisions while increasing the stability of their system.

First, we highlight how different perturbation types have different effects on the modules forming the RAG system and its end-to-end performance. Our experiments showed that certain perturbations and dataset combinations lead to more sensitivity on a specific RAG component. Therefore, we recommend that practitioners use our analysis framework on their we recommend that practitioners use our analysis framework on their own data for a better diagnosis of sensitivity in their pipeline.

We acknowledge that the robustness of the generators is generally assessed in the closed-book setting without their use in a RAG pipeline. However, as our results show, certain query perturbations affect the response generation differently when documents are presented in the context window of the generator. Therefore, we recommend that practitioners assess the robustness of the response generation in their systems, especially in an oracle setting, as this setting estimates an upper bound for the system.

Furthermore, there is an active field of study investigating the training of retrieval augmented generation systems where the retriever and the LLM are jointly trained \citep{radit}. These systems benefit from training by becoming more robust against irrelevant contexts when generating the answer. However, the robustness against the query variations in the context of joint training is still underexplored. Our findings can be integrated into these training regimes to develop robustness-aware end-to-end systems that are stable against query variations.

Lastly, as many methods of query disambiguation and expansion show, the query transformations help increase the performance of RAG systems while introducing extra computational overhead. Since these methods are greatly dependent on the initial query provided to the system and could be employed for different modules of the pipeline, we believe that our decoupling analysis could help practitioners identify the sensitive modules in their pipeline to employ these methods more efficiently for different perturbation types. Therefore, we recommend using our provided analysis methodology to test the RAG system before employing query transformations steps into the pipeline.

\section{Conclusion}

In this work, we highlight a key issue of the retrieval augmented generation pipelines, namely their sensitivity to query variations. We perform extensive experiments on three question answering datasets and twelve RAG pipelines which span four retrievers and three LLM generators. Our experiments probing these pipelines with five perturbation types show that slight variations in the input queries can result in significant performance discrepancies.

Our analysis framework enables the investigation of RAG module sensitivity to query perturbations jointly and in isolation. Using this framework, we provide practical insights and recommendations for the development of RAG systems. We hope that our work brings greater attention to the importance of robustness research at the query level while contributing to the development of future robustness-aware retrieval augmented generation pipelines.

\section*{Limitations}

Due to computational and time limitations, our experiments are constrained to have a limited context window length, number of output tokens generated, and maximum length defined for each text passage inserted into the inference of the LLMs. We acknowledge the limitations of our system and provide a comparative analysis between the pipeline combinations. Hence, the exploration of the hyperparameter space to formulate optimal pipeline configurations remains a potential avenue for future research. This search also includes the prompt tuning for sample generation and question answering. In future work, prompts will be further tuned to meet the characteristics of datasets better.

Furthermore, we kept our pipeline simple to provide researchers with a framework to evaluate their own system. With the same aim and to reflect the scenario where there is limited computational power and high-quality annotated data, we chose to use retrievers and LLMs directly without applying fine-tuning. This decision also entailed the exclusion of rerankers as these models also rely on the performance of retrievers to return document sets with larger set sizes. Our analysis shows that the perturbations are still effective on larger document set sizes as shown in Figure \ref{prec_vs_recall} and \ref{fig: remaining_retriever_results}, therefore we leave the analysis of the rerankers to a future study.

Moreover, the role of the ranking of the document sets returned by the retriever with respect to the perturbations is left to a future study. We hope that by integrating metrics that concentrate more on the ranking aspects of the retrieval (e.g. MRR and nDCG) into our analysis framework, practitioners can assess the sensitivity of their pipelines with a focus on this particular aspect.

Lastly, potential mitigation strategies aiming to increase the robustness of the modules such as fine-tuning of the retrieval augmented generation components on the perturbed sample-answer pairs, or including perturbed samples into the end-to-end joint training of retrievers and LLMs for robustness aware question answering systems are not discussed within the scope of the analysis of this work. This also includes the analysis of another category of methods in relation to query perturbations, namely the query transformations. The robustness of these methods and their effect within the scope of RAG pipelines when faced with various input variations are not addressed in this study. Lastly, while the investigation of LLM internal representations under different perturbations is included in our analysis, its dedicated in-depth analysis is still of interest as a promising research direction. We recognize that the absence of these points in our analysis is a limitation and will address these approaches as a part of our future study.


\bibliography{custom}

\begin{thebibliography}{44}
\providecommand{\natexlab}[1]{#1}

\bibitem[{Arabzadeh et~al.(2023)Arabzadeh, Hamidi~Rad, Khodabakhsh, and Bagheri}]{arabzadeh2023noisy}
Negar Arabzadeh, Radin Hamidi~Rad, Maryam Khodabakhsh, and Ebrahim Bagheri. 2023.
\newblock Noisy perturbations for estimating query difficulty in dense retrievers.
\newblock In \emph{Proceedings of the 32nd ACM International Conference on Information and Knowledge Management}, pages 3722--3727.

\bibitem[{Chen et~al.(2023)Chen, Lin, Han, and Sun}]{rag_robustness1}
Jiawei Chen, Hongyu Lin, Xianpei Han, and Le~Sun. 2023.
\newblock \href {https://arxiv.org/abs/2309.01431} {Benchmarking large language models in retrieval-augmented generation}.
\newblock \emph{Preprint}, arXiv:2309.01431.

\bibitem[{Fang et~al.(2024)Fang, Bai, Ni, Yang, Chen, and Xu}]{rag_robustness5}
Feiteng Fang, Yuelin Bai, Shiwen Ni, Min Yang, Xiaojun Chen, and Ruifeng Xu. 2024.
\newblock \href {https://arxiv.org/abs/2405.20978} {Enhancing noise robustness of retrieval-augmented language models with adaptive adversarial training}.
\newblock \emph{Preprint}, arXiv:2405.20978.

\bibitem[{Grattafiori et~al.(2024)Grattafiori, Dubey, Jauhri, Pandey, Kadian, Al-Dahle, Letman, Mathur, Schelten, Vaughan, Yang, Fan, Goyal, Hartshorn, Yang, Mitra, Sravankumar, Korenev, Hinsvark, Rao, Zhang, Rodriguez, Gregerson, Spataru, Roziere, Biron, Tang, Chern, Caucheteux, Nayak, Bi, Marra, McConnell, Keller, Touret, Wu, Wong, Ferrer, Nikolaidis, Allonsius, Song, Pintz, Livshits, Wyatt, Esiobu, Choudhary, Mahajan, Garcia-Olano, Perino, Hupkes, Lakomkin, AlBadawy, Lobanova, Dinan, Smith, Radenovic, Guzmán, Zhang, Synnaeve, Lee, Anderson, Thattai, Nail, Mialon, Pang, Cucurell, Nguyen, Korevaar, Xu, Touvron, Zarov, Ibarra, Kloumann, Misra, Evtimov, Zhang, Copet, Lee, Geffert, Vranes, Park, Mahadeokar, Shah, van~der Linde, Billock, Hong, Lee, Fu, Chi, Huang, Liu, Wang, Yu, Bitton, Spisak, Park, Rocca, Johnstun, Saxe, Jia, Alwala, Prasad, Upasani, Plawiak, Li, Heafield, Stone, El-Arini, Iyer, Malik, Chiu, Bhalla, Lakhotia, Rantala-Yeary, van~der Maaten, Chen, Tan, Jenkins, Martin, Madaan, Malo, Blecher,
  Landzaat, de~Oliveira, Muzzi, Pasupuleti, Singh, Paluri, Kardas, Tsimpoukelli, Oldham, Rita, Pavlova, Kambadur, Lewis, Si, Singh, Hassan, Goyal, Torabi, Bashlykov, Bogoychev, Chatterji, Zhang, Duchenne, Çelebi, Alrassy, Zhang, Li, Vasic, Weng, Bhargava, Dubal, Krishnan, Koura, Xu, He, Dong, Srinivasan, Ganapathy, Calderer, Cabral, Stojnic, Raileanu, Maheswari, Girdhar, Patel, Sauvestre, Polidoro, Sumbaly, Taylor, Silva, Hou, Wang, Hosseini, Chennabasappa, Singh, Bell, Kim, Edunov, Nie, Narang, Raparthy, Shen, Wan, Bhosale, Zhang, Vandenhende, Batra, Whitman, Sootla, Collot, Gururangan, Borodinsky, Herman, Fowler, Sheasha, Georgiou, Scialom, Speckbacher, Mihaylov, Xiao, Karn, Goswami, Gupta, Ramanathan, Kerkez, Gonguet, Do, Vogeti, Albiero, Petrovic, Chu, Xiong, Fu, Meers, Martinet, Wang, Wang, Tan, Xia, Xie, Jia, Wang, Goldschlag, Gaur, Babaei, Wen, Song, Zhang, Li, Mao, Coudert, Yan, Chen, Papakipos, Singh, Srivastava, Jain, Kelsey, Shajnfeld, Gangidi, Victoria, Goldstand, Menon, Sharma, Boesenberg,
  Baevski, Feinstein, Kallet, Sangani, Teo, Yunus, Lupu, Alvarado, Caples, Gu, Ho, Poulton, Ryan, Ramchandani, Dong, Franco, Goyal, Saraf, Chowdhury, Gabriel, Bharambe, Eisenman, Yazdan, James, Maurer, Leonhardi, Huang, Loyd, Paola, Paranjape, Liu, Wu, Ni, Hancock, Wasti, Spence, Stojkovic, Gamido, Montalvo, Parker, Burton, Mejia, Liu, Wang, Kim, Zhou, Hu, Chu, Cai, Tindal, Feichtenhofer, Gao, Civin, Beaty, Kreymer, Li, Adkins, Xu, Testuggine, David, Parikh, Liskovich, Foss, Wang, Le, Holland, Dowling, Jamil, Montgomery, Presani, Hahn, Wood, Le, Brinkman, Arcaute, Dunbar, Smothers, Sun, Kreuk, Tian, Kokkinos, Ozgenel, Caggioni, Kanayet, Seide, Florez, Schwarz, Badeer, Swee, Halpern, Herman, Sizov, Guangyi, Zhang, Lakshminarayanan, Inan, Shojanazeri, Zou, Wang, Zha, Habeeb, Rudolph, Suk, Aspegren, Goldman, Zhan, Damlaj, Molybog, Tufanov, Leontiadis, Veliche, Gat, Weissman, Geboski, Kohli, Lam, Asher, Gaya, Marcus, Tang, Chan, Zhen, Reizenstein, Teboul, Zhong, Jin, Yang, Cummings, Carvill, Shepard, McPhie,
  Torres, Ginsburg, Wang, Wu, U, Saxena, Khandelwal, Zand, Matosich, Veeraraghavan, Michelena, Li, Jagadeesh, Huang, Chawla, Huang, Chen, Garg, A, Silva, Bell, Zhang, Guo, Yu, Moshkovich, Wehrstedt, Khabsa, Avalani, Bhatt, Mankus, Hasson, Lennie, Reso, Groshev, Naumov, Lathi, Keneally, Liu, Seltzer, Valko, Restrepo, Patel, Vyatskov, Samvelyan, Clark, Macey, Wang, Hermoso, Metanat, Rastegari, Bansal, Santhanam, Parks, White, Bawa, Singhal, Egebo, Usunier, Mehta, Laptev, Dong, Cheng, Chernoguz, Hart, Salpekar, Kalinli, Kent, Parekh, Saab, Balaji, Rittner, Bontrager, Roux, Dollar, Zvyagina, Ratanchandani, Yuvraj, Liang, Alao, Rodriguez, Ayub, Murthy, Nayani, Mitra, Parthasarathy, Li, Hogan, Battey, Wang, Howes, Rinott, Mehta, Siby, Bondu, Datta, Chugh, Hunt, Dhillon, Sidorov, Pan, Mahajan, Verma, Yamamoto, Ramaswamy, Lindsay, Lindsay, Feng, Lin, Zha, Patil, Shankar, Zhang, Zhang, Wang, Agarwal, Sajuyigbe, Chintala, Max, Chen, Kehoe, Satterfield, Govindaprasad, Gupta, Deng, Cho, Virk, Subramanian, Choudhury,
  Goldman, Remez, Glaser, Best, Koehler, Robinson, Li, Zhang, Matthews, Chou, Shaked, Vontimitta, Ajayi, Montanez, Mohan, Kumar, Mangla, Ionescu, Poenaru, Mihailescu, Ivanov, Li, Wang, Jiang, Bouaziz, Constable, Tang, Wu, Wang, Wu, Gao, Kleinman, Chen, Hu, Jia, Qi, Li, Zhang, Zhang, Adi, Nam, Yu, Wang, Zhao, Hao, Qian, Li, He, Rait, DeVito, Rosnbrick, Wen, Yang, Zhao, and Ma}]{llama31}
Aaron Grattafiori, Abhimanyu Dubey, Abhinav Jauhri, Abhinav Pandey, Abhishek Kadian, Ahmad Al-Dahle, Aiesha Letman, Akhil Mathur, Alan Schelten, Alex Vaughan, Amy Yang, Angela Fan, Anirudh Goyal, Anthony Hartshorn, Aobo Yang, Archi Mitra, Archie Sravankumar, Artem Korenev, Arthur Hinsvark, and 542 others. 2024.
\newblock \href {https://arxiv.org/abs/2407.21783} {The llama 3 herd of models}.
\newblock \emph{Preprint}, arXiv:2407.21783.

\bibitem[{Han et~al.(2023)Han, Qi, Zhang, Liu, Burger, Wang, Huang, Xiang, and Roth}]{han2023robustqa}
Rujun Han, Peng Qi, Yuhao Zhang, Lan Liu, Juliette Burger, William~Yang Wang, Zhiheng Huang, Bing Xiang, and Dan Roth. 2023.
\newblock Robustqa: Benchmarking the robustness of domain adaptation for open-domain question answering.
\newblock In \emph{Findings of the Association for Computational Linguistics: ACL 2023}, pages 4294--4311.

\bibitem[{Hsia et~al.(2024)Hsia, Shaikh, Wang, and Neubig}]{ragged}
Jennifer Hsia, Afreen Shaikh, Zhiruo Wang, and Graham Neubig. 2024.
\newblock \href {https://arxiv.org/abs/2403.09040} {Ragged: Towards informed design of retrieval augmented generation systems}.
\newblock \emph{Preprint}, arXiv:2403.09040.

\bibitem[{Hu et~al.(2024)Hu, Wang, Shu, Helen, Paik, and Zhu}]{rag_robustness2}
Zhibo Hu, Chen Wang, Yanfeng Shu, Helen, Paik, and Liming Zhu. 2024.
\newblock \href {https://arxiv.org/abs/2402.07179} {Prompt perturbation in retrieval-augmented generation based large language models}.
\newblock \emph{Preprint}, arXiv:2402.07179.

\bibitem[{Huang et~al.(2023)Huang, Tao, An, Zhang, Jiang, Chen, Wu, and Feng}]{domspecifichallucination}
Quzhe Huang, Mingxu Tao, Zhenwei An, Chen Zhang, Cong Jiang, Zhibin Chen, Zirui Wu, and Yansong Feng. 2023.
\newblock \href {https://doi.org/10.48550/ARXIV.2305.15062} {Lawyer llama technical report}.
\newblock \emph{CoRR}, abs/2305.15062.

\bibitem[{Izacard et~al.(2022)Izacard, Caron, Hosseini, Riedel, Bojanowski, Joulin, and Grave}]{contriever}
Gautier Izacard, Mathilde Caron, Lucas Hosseini, Sebastian Riedel, Piotr Bojanowski, Armand Joulin, and Edouard Grave. 2022.
\newblock \href {https://arxiv.org/abs/2112.09118} {Unsupervised dense information retrieval with contrastive learning}.
\newblock \emph{Preprint}, arXiv:2112.09118.

\bibitem[{Jiang et~al.(2023)Jiang, Sablayrolles, Mensch, Bamford, Chaplot, de~las Casas, Bressand, Lengyel, Lample, Saulnier, Lavaud, Lachaux, Stock, Scao, Lavril, Wang, Lacroix, and Sayed}]{mistral7b}
Albert~Q. Jiang, Alexandre Sablayrolles, Arthur Mensch, Chris Bamford, Devendra~Singh Chaplot, Diego de~las Casas, Florian Bressand, Gianna Lengyel, Guillaume Lample, Lucile Saulnier, Lélio~Renard Lavaud, Marie-Anne Lachaux, Pierre Stock, Teven~Le Scao, Thibaut Lavril, Thomas Wang, Timothée Lacroix, and William~El Sayed. 2023.
\newblock \href {https://arxiv.org/abs/2310.06825} {Mistral 7b}.
\newblock \emph{Preprint}, arXiv:2310.06825.

\bibitem[{Kwiatkowski et~al.(2019{\natexlab{a}})Kwiatkowski, Palomaki, Redfield, Collins, Parikh, Alberti, Epstein, Polosukhin, Devlin, Lee, Toutanova, Jones, Kelcey, Chang, Dai, Uszkoreit, Le, and Petrov}]{dataset_nq}
Tom Kwiatkowski, Jennimaria Palomaki, Olivia Redfield, Michael Collins, Ankur Parikh, Chris Alberti, Danielle Epstein, Illia Polosukhin, Jacob Devlin, Kenton Lee, Kristina Toutanova, Llion Jones, Matthew Kelcey, Ming-Wei Chang, Andrew~M. Dai, Jakob Uszkoreit, Quoc Le, and Slav Petrov. 2019{\natexlab{a}}.
\newblock \href {https://doi.org/10.1162/tacl_a_00276} {Natural questions: A benchmark for question answering research}.
\newblock \emph{Transactions of the Association for Computational Linguistics}, 7:452--466.

\bibitem[{Kwiatkowski et~al.(2019{\natexlab{b}})Kwiatkowski, Palomaki, Redfield, Collins, Parikh, Alberti, Epstein, Polosukhin, Devlin, Lee, Toutanova, Jones, Kelcey, Chang, Dai, Uszkoreit, Le, and Petrov}]{nqdataset}
Tom Kwiatkowski, Jennimaria Palomaki, Olivia Redfield, Michael Collins, Ankur Parikh, Chris Alberti, Danielle Epstein, Illia Polosukhin, Jacob Devlin, Kenton Lee, Kristina Toutanova, Llion Jones, Matthew Kelcey, Ming-Wei Chang, Andrew~M. Dai, Jakob Uszkoreit, Quoc Le, and Slav Petrov. 2019{\natexlab{b}}.
\newblock \href {https://doi.org/10.1162/tacl_a_00276} {Natural questions: A benchmark for question answering research}.
\newblock \emph{Transactions of the Association for Computational Linguistics}, 7:452--466.

\bibitem[{Kwon et~al.(2023)Kwon, Li, Zhuang, Sheng, Zheng, Yu, Gonzalez, Zhang, and Stoica}]{vllm}
Woosuk Kwon, Zhuohan Li, Siyuan Zhuang, Ying Sheng, Lianmin Zheng, Cody~Hao Yu, Joseph~E. Gonzalez, Hao Zhang, and Ion Stoica. 2023.
\newblock Efficient memory management for large language model serving with pagedattention.
\newblock In \emph{Proceedings of the ACM SIGOPS 29th Symposium on Operating Systems Principles}.

\bibitem[{Lewis et~al.(2020)Lewis, Perez, Piktus, Petroni, Karpukhin, Goyal, K{\"u}ttler, Lewis, Yih, Rockt{\"a}schel et~al.}]{lewis2020retrieval}
Patrick Lewis, Ethan Perez, Aleksandra Piktus, Fabio Petroni, Vladimir Karpukhin, Naman Goyal, Heinrich K{\"u}ttler, Mike Lewis, Wen-tau Yih, Tim Rockt{\"a}schel, and 1 others. 2020.
\newblock Retrieval-augmented generation for knowledge-intensive nlp tasks.
\newblock \emph{Advances in Neural Information Processing Systems}, 33:9459--9474.

\bibitem[{Lin et~al.(2021)Lin, Ma, Lin, Yang, Pradeep, and Nogueira}]{pyserini}
Jimmy Lin, Xueguang Ma, Sheng-Chieh Lin, Jheng-Hong Yang, Ronak Pradeep, and Rodrigo Nogueira. 2021.
\newblock \href {https://arxiv.org/abs/2102.10073} {Pyserini: An easy-to-use python toolkit to support replicable ir research with sparse and dense representations}.
\newblock \emph{Preprint}, arXiv:2102.10073.

\bibitem[{Lin et~al.(2024)Lin, Chen, Chen, Shi, Lomeli, James, Rodriguez, Kahn, Szilvasy, Lewis, Zettlemoyer, and Yih}]{radit}
Xi~Victoria Lin, Xilun Chen, Mingda Chen, Weijia Shi, Maria Lomeli, Rich James, Pedro Rodriguez, Jacob Kahn, Gergely Szilvasy, Mike Lewis, Luke Zettlemoyer, and Scott Yih. 2024.
\newblock \href {https://arxiv.org/abs/2310.01352} {Ra-dit: Retrieval-augmented dual instruction tuning}.
\newblock \emph{Preprint}, arXiv:2310.01352.

\bibitem[{Liu et~al.(2023)Liu, Zhang, Guo, Chen, and Cheng}]{liu2023robustness}
Yu-An Liu, Ruqing Zhang, Jiafeng Guo, Wei Chen, and Xueqi Cheng. 2023.
\newblock On the robustness of generative retrieval models: An out-of-distribution perspective.
\newblock \emph{arXiv preprint arXiv:2306.12756}.

\bibitem[{Liu et~al.(2024)Liu, Zhang, Guo, de~Rijke, Fan, and Cheng}]{neuralIR_robustness}
Yu-An Liu, Ruqing Zhang, Jiafeng Guo, Maarten de~Rijke, Yixing Fan, and Xueqi Cheng. 2024.
\newblock \href {https://arxiv.org/abs/2407.06992} {Robust neural information retrieval: An adversarial and out-of-distribution perspective}.
\newblock \emph{Preprint}, arXiv:2407.06992.

\bibitem[{Morris et~al.(2020)Morris, Lifland, Yoo, Grigsby, Jin, and Qi}]{textattack}
John~X. Morris, Eli Lifland, Jin~Yong Yoo, Jake Grigsby, Di~Jin, and Yanjun Qi. 2020.
\newblock \href {https://arxiv.org/abs/2005.05909} {Textattack: A framework for adversarial attacks, data augmentation, and adversarial training in nlp}.
\newblock \emph{Preprint}, arXiv:2005.05909.

\bibitem[{Penha et~al.(2022)Penha, Câmara, and Hauff}]{oodqueryrobustness1}
Gustavo Penha, Arthur Câmara, and Claudia Hauff. 2022.
\newblock \href {https://arxiv.org/abs/2111.13057} {Evaluating the robustness of retrieval pipelines with query variation generators}.
\newblock \emph{Preprint}, arXiv:2111.13057.

\bibitem[{Radford et~al.(2019)Radford, Wu, Child, Luan, Amodei, Sutskever et~al.}]{gpt2large}
Alec Radford, Jeffrey Wu, Rewon Child, David Luan, Dario Amodei, Ilya Sutskever, and 1 others. 2019.
\newblock Language models are unsupervised multitask learners.
\newblock \emph{OpenAI blog}, 1(8):9.

\bibitem[{Ram et~al.(2023)Ram, Levine, Dalmedigos, Muhlgay, Shashua, Leyton-Brown, and Shoham}]{ram2023context}
Ori Ram, Yoav Levine, Itay Dalmedigos, Dor Muhlgay, Amnon Shashua, Kevin Leyton-Brown, and Yoav Shoham. 2023.
\newblock In-context retrieval-augmented language models.
\newblock \emph{Transactions of the Association for Computational Linguistics}, 11:1316--1331.

\bibitem[{Rau et~al.(2024)Rau, Déjean, Chirkova, Formal, Wang, Nikoulina, and Clinchant}]{bergen}
David Rau, Hervé Déjean, Nadezhda Chirkova, Thibault Formal, Shuai Wang, Vassilina Nikoulina, and Stéphane Clinchant. 2024.
\newblock \href {https://arxiv.org/abs/2407.01102} {Bergen: A benchmarking library for retrieval-augmented generation}.
\newblock \emph{Preprint}, arXiv:2407.01102.

\bibitem[{Robertson et~al.(1995)Robertson, Walker, Jones, Hancock-Beaulieu, Gatford et~al.}]{BM25}
Stephen~E Robertson, Steve Walker, Susan Jones, Micheline~M Hancock-Beaulieu, Mike Gatford, and 1 others. 1995.
\newblock Okapi at trec-3.
\newblock \emph{Nist Special Publication Sp}, 109:109.

\bibitem[{Shen et~al.(2024)Shen, Blloshmi, Zhu, Pei, and Zhang}]{shen2024assessing}
Xiaoyu Shen, Rexhina Blloshmi, Dawei Zhu, Jiahuan Pei, and Wei Zhang. 2024.
\newblock Assessing “implicit” retrieval robustness of large language models.
\newblock In \emph{Proceedings of the 2024 Conference on Empirical Methods in Natural Language Processing}, pages 8988--9003.

\bibitem[{Shi et~al.(2023)Shi, Chen, Misra, Scales, Dohan, Chi, Schärli, and Zhou}]{llm_robustness1}
Freda Shi, Xinyun Chen, Kanishka Misra, Nathan Scales, David Dohan, Ed~Chi, Nathanael Schärli, and Denny Zhou. 2023.
\newblock \href {https://arxiv.org/abs/2302.00093} {Large language models can be easily distracted by irrelevant context}.
\newblock \emph{Preprint}, arXiv:2302.00093.

\bibitem[{Sidiropoulos and Kanoulas(2022)}]{sidiropoulos2022analysing}
Georgios Sidiropoulos and Evangelos Kanoulas. 2022.
\newblock Analysing the robustness of dual encoders for dense retrieval against misspellings.
\newblock In \emph{Proceedings of the 45th International ACM SIGIR Conference on Research and Development in Information Retrieval}, pages 2132--2136.

\bibitem[{Sun et~al.(2023)Sun, Shaib, and Wallace}]{para2}
Jiuding Sun, Chantal Shaib, and Byron~C. Wallace. 2023.
\newblock \href {https://arxiv.org/abs/2306.11270} {Evaluating the zero-shot robustness of instruction-tuned language models}.
\newblock \emph{Preprint}, arXiv:2306.11270.

\bibitem[{Team(2024)}]{qwen2.5}
Qwen Team. 2024.
\newblock \href {https://qwenlm.github.io/blog/qwen2.5/} {Qwen2.5: A party of foundation models}.

\bibitem[{Thakur et~al.(2021)Thakur, Reimers, Rücklé, Srivastava, and Gurevych}]{BEIR}
Nandan Thakur, Nils Reimers, Andreas Rücklé, Abhishek Srivastava, and Iryna Gurevych. 2021.
\newblock \href {https://arxiv.org/abs/2104.08663} {Beir: A heterogenous benchmark for zero-shot evaluation of information retrieval models}.
\newblock \emph{Preprint}, arXiv:2104.08663.

\bibitem[{Tsatsaronis et~al.(2015)Tsatsaronis, Balikas, Malakasiotis, Partalas, Zschunke, Alvers, Weißenborn, Krithara, Petridis, Polychronopoulos, Almirantis, Pavlopoulos, Baskiotis, Gallinari, Artieres, Ngonga~Ngomo, Heino, Gaussier, Barrio-Alvers, and Paliouras}]{dataset_bioasq}
George Tsatsaronis, Georgios Balikas, Prodromos Malakasiotis, Ioannis Partalas, Matthias Zschunke, Michael Alvers, Dirk Weißenborn, Anastasia Krithara, Sergios Petridis, Dimitris Polychronopoulos, Yannis Almirantis, John Pavlopoulos, Nicolas Baskiotis, Patrick Gallinari, Thierry Artieres, Axel-Cyrille Ngonga~Ngomo, Norman Heino, Eric Gaussier, Liliana Barrio-Alvers, and Georgios Paliouras. 2015.
\newblock \href {https://doi.org/10.1186/s12859-015-0564-6} {An overview of the bioasq large-scale biomedical semantic indexing and question answering competition}.
\newblock \emph{BMC Bioinformatics}, 16:138.

\bibitem[{Wang et~al.(2024)Wang, Yang, Huang, Yang, Majumder, and Wei}]{multie5base}
Liang Wang, Nan Yang, Xiaolong Huang, Linjun Yang, Rangan Majumder, and Furu Wei. 2024.
\newblock Multilingual e5 text embeddings: A technical report.
\newblock \emph{arXiv preprint arXiv:2402.05672}.

\bibitem[{Xiang et~al.(2024)Xiang, Wu, Zhong, Wagner, Chen, and Mittal}]{rag_robustness4}
Chong Xiang, Tong Wu, Zexuan Zhong, David Wagner, Danqi Chen, and Prateek Mittal. 2024.
\newblock \href {https://arxiv.org/abs/2405.15556} {Certifiably robust rag against retrieval corruption}.
\newblock \emph{Preprint}, arXiv:2405.15556.

\bibitem[{Xiao et~al.(2024)Xiao, Liu, Zhang, Muennighoff, Lian, and Nie}]{bgebase}
Shitao Xiao, Zheng Liu, Peitian Zhang, Niklas Muennighoff, Defu Lian, and Jian-Yun Nie. 2024.
\newblock \href {https://arxiv.org/abs/2309.07597} {C-pack: Packed resources for general chinese embeddings}.
\newblock \emph{Preprint}, arXiv:2309.07597.

\bibitem[{Xie et~al.(2024)Xie, Zhang, Duan, Zhang, and Huang}]{textattackref3}
Guicai Xie, Ke~Zhang, Lei Duan, Wei Zhang, and Zeqian Huang. 2024.
\newblock \href {https://aclanthology.org/2024.lrec-main.1470/} {Typos correction training against misspellings from text-to-text transformers}.
\newblock In \emph{Proceedings of the 2024 Joint International Conference on Computational Linguistics, Language Resources and Evaluation (LREC-COLING 2024)}, pages 16907--16918, Torino, Italia. ELRA and ICCL.

\bibitem[{Yang et~al.(2018)Yang, Qi, Zhang, Bengio, Cohen, Salakhutdinov, and Manning}]{dataset_hotpotqa}
Zhilin Yang, Peng Qi, Saizheng Zhang, Yoshua Bengio, William Cohen, Ruslan Salakhutdinov, and Christopher~D. Manning. 2018.
\newblock \href {https://doi.org/10.18653/v1/D18-1259} {{H}otpot{QA}: A dataset for diverse, explainable multi-hop question answering}.
\newblock In \emph{Proceedings of the 2018 Conference on Empirical Methods in Natural Language Processing}, pages 2369--2380, Brussels, Belgium. Association for Computational Linguistics.

\bibitem[{Yoran et~al.(2024)Yoran, Wolfson, Ram, and Berant}]{rag_robustness3}
Ori Yoran, Tomer Wolfson, Ori Ram, and Jonathan Berant. 2024.
\newblock \href {https://arxiv.org/abs/2310.01558} {Making retrieval-augmented language models robust to irrelevant context}.
\newblock \emph{Preprint}, arXiv:2310.01558.

\bibitem[{Zhang et~al.(2025)Zhang, Tang, Ruan, Huang, Khastgir, Jennings, and Zhao}]{textattackref2}
Yi~Zhang, Yun Tang, Wenjie Ruan, Xiaowei Huang, Siddartha Khastgir, Paul Jennings, and Xingyu Zhao. 2025.
\newblock Protip: Probabilistic robustness verification on text-to-image diffusion models against stochastic perturbation.
\newblock In \emph{European Conference on Computer Vision}, pages 455--472. Springer.

\bibitem[{Zhao et~al.(2024{\natexlab{a}})Zhao, Chen, Yang, Liu, Deng, Cai, Wang, Yin, and Du}]{llmtransp}
Haiyan Zhao, Hanjie Chen, Fan Yang, Ninghao Liu, Huiqi Deng, Hengyi Cai, Shuaiqiang Wang, Dawei Yin, and Mengnan Du. 2024{\natexlab{a}}.
\newblock \href {https://doi.org/10.1145/3639372} {Explainability for large language models: A survey}.
\newblock \emph{ACM Trans. Intell. Syst. Technol.}, 15(2).

\bibitem[{Zhao et~al.(2024{\natexlab{b}})Zhao, Yan, Sun, Xing, Wang, Meng, Cheng, Ren, and Yin}]{para1}
Yukun Zhao, Lingyong Yan, Weiwei Sun, Guoliang Xing, Shuaiqiang Wang, Chong Meng, Zhicong Cheng, Zhaochun Ren, and Dawei Yin. 2024{\natexlab{b}}.
\newblock Improving the robustness of large language models via consistency alignment.
\newblock \emph{arXiv preprint arXiv:2403.14221}.

\bibitem[{Zhao et~al.(2024{\natexlab{c}})Zhao, Yan, Sun, Xing, Wang, Meng, Cheng, Ren, and Yin}]{para3}
Yukun Zhao, Lingyong Yan, Weiwei Sun, Guoliang Xing, Shuaiqiang Wang, Chong Meng, Zhicong Cheng, Zhaochun Ren, and Dawei Yin. 2024{\natexlab{c}}.
\newblock \href {https://arxiv.org/abs/2403.14221} {Improving the robustness of large language models via consistency alignment}.
\newblock \emph{Preprint}, arXiv:2403.14221.

\bibitem[{Zhu et~al.(2024{\natexlab{a}})Zhu, Wang, Zhou, Wang, Chen, Wang, Yang, Ye, Zhang, Gong, and Xie}]{llm_robustness2}
Kaijie Zhu, Jindong Wang, Jiaheng Zhou, Zichen Wang, Hao Chen, Yidong Wang, Linyi Yang, Wei Ye, Yue Zhang, Neil~Zhenqiang Gong, and Xing Xie. 2024{\natexlab{a}}.
\newblock \href {https://arxiv.org/abs/2306.04528} {Promptrobust: Towards evaluating the robustness of large language models on adversarial prompts}.
\newblock \emph{Preprint}, arXiv:2306.04528.

\bibitem[{Zhu et~al.(2024{\natexlab{b}})Zhu, Zhao, Chen, Wang, and Xie}]{promptbench}
Kaijie Zhu, Qinlin Zhao, Hao Chen, Jindong Wang, and Xing Xie. 2024{\natexlab{b}}.
\newblock Promptbench: A unified library for evaluation of large language models.
\newblock \emph{Journal of Machine Learning Research}, 25(254):1--22.

\bibitem[{Zhuang and Zuccon(2022)}]{charbert}
Shengyao Zhuang and Guido Zuccon. 2022.
\newblock \href {https://doi.org/10.1145/3477495.3531951} {Characterbert and self-teaching for improving the robustness of dense retrievers on queries with typos}.
\newblock In \emph{Proceedings of the 45th International ACM SIGIR Conference on Research and Development in Information Retrieval}, SIGIR ’22, page 1444–1454. ACM.

\end{thebibliography}

\clearpage
\appendix

\section{Appendix}
\label{sec:appendix}

In this appendix we provide more details of the data preparation, perturbation and experiments runs. 

\subsection{Datasets}

In this study, the experiments are performed on three datasets that are included from the BEIR benchmark: HotpotQA, Natural Questions and BioASQ. For all of the datasets, we incorporated the corpora defined within the BEIR benchmark and used samples from the test split of the datasets.

\textbf{HotpotQA} dataset is multi-hop question answering dataset that uses Wikipedia as knowledge base. This dataset requires system to retrieve all the reference text passages and generators in the system to reason over them  \citep{dataset_hotpotqa}.

\textbf{Natural Questions (NQ)} dataset is single-hop question answering dataset consisting of generic questions and named as "natural" since the collected questions are collected from the real user queries submitted to the Google Search Engine \citep{dataset_nq}. 

\textbf{BioASQ} dataset is a biomedical question-answering dataset in English that uses articles from PubMed as its corpus. BEIR benchmark uses the Training v.2020 data for task 9a as corpus while using the test data from the task 8b as queries. Further detail on the number of samples and corpus sizes as well dataset characteristics could be seen in Table \ref{sample-table}  \citep{dataset_bioasq}.

\begin{figure*}[t]
  \includegraphics[width=1\linewidth]{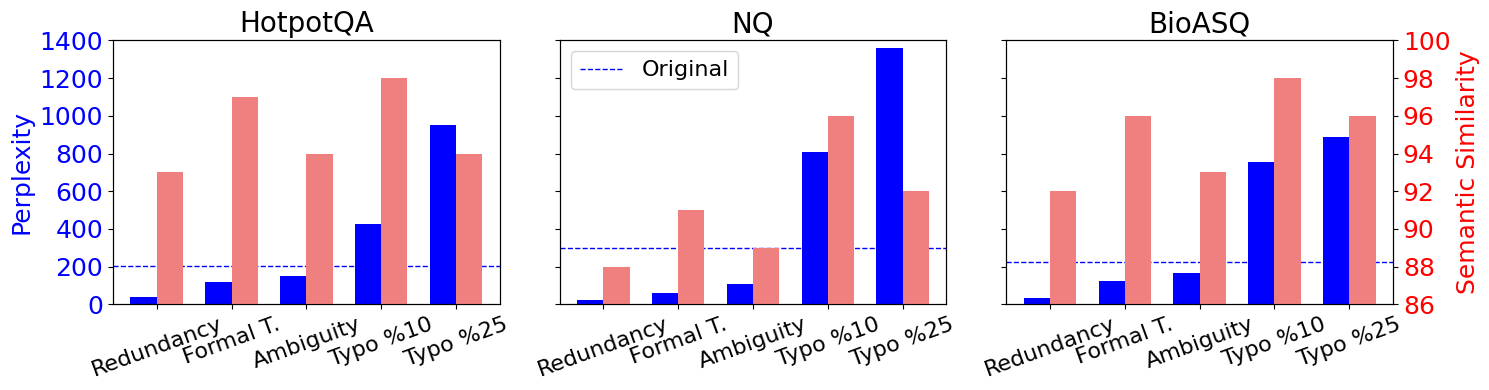}
  \caption{Perplexity and semantic similarity of the generated samples for different perturbations and datasets.}
  \label{sample_quality_persim}
\end{figure*}
\subsection{Automated Sample Generation}\label{app_aut_sample_gen}

Transforming textual input using large language models is a widely used technique in natural langueg processing community.For instance, the \citep{para1,para2, para3} used large language models to generate paraphrases of the textual inputs. Following the previous work, we also used GPT4o to automatically generate the perturbed samples. The prompts used to generate perturbed samples for redundancy, formal tone and ambiguity insertion cases can be found in Table \ref{tab: prompts}.

\begin{table}[t]
  \begin{tabular}{p{1\linewidth}}
    \hline
    \textbf{Redundancy}  \\
    \hline
    """Paraphrase the input text \{output\_per\_sample\} times by inserting related redundant knowledge into the input text. Do not insert any information that will answer the question directly.\\\\

Separate the output text samples by single \symbol{92}n between them. Do not output anything else and do not answer the question but only paraphrase it.

Input text: \{input\_str\}\\
Output:\symbol{92}n\symbol{92}n
"""\\
    \hline
    \textbf{Formality} \\ 
 \hline
 """Paraphrase the input text \{output\_per\_sample\} times in a more formal tone.\\\\

Separate the output text samples by single \symbol{92}n between them. Do not output anything else and do not answer the question but only paraphrase it.

Input text: \{input\_str\}\\
Output:\symbol{92}n\symbol{92}n
"""\\
\hline
\textbf{Ambiguity} \\
 \hline
 """Paraphrase the text below \{output\_per\_sample\}  times while making it unclear to answer by introducing ambiguity to the text.\\
Separate the output text samples by single \symbol{92}n between them. Do not output anything else and do not answer the question but only paraphrase it.\\
Input text: \{input\_str\}\\
Output:\symbol{92}n\symbol{92}n
"""\\
  \end{tabular}
  \caption{\label{tab: prompts} Prompts used to generate perturbed samples.}
\end{table}

To assess the quality of the generated samples, we checked the perplexity and the semantic similarity values corresponding to different perturbation types which are shown in Figure \ref{sample_quality_persim}. For the perplexity calculations, we used GPT2-Large \citep{gpt2large} calculate the perplexity values for each sample corresponding to a perturbation and reported the mean perplexity values. The average perplexity of the original, i.e. non-perturbed, samples are reported with the dashed line on Figure \ref{sample_quality_persim} for each dataset. The results showed that the samples perturbed via prompting have less perplexity when compared to the original samples while the typo insertions result in samples more perplexing to the models. Based on our analysis, we showed that the performance degradations do not stem from the naturalness of the samples to the large language models.

Further, we calculated the semantic similarity of the samples to the original ones by embedding the samples into a vector space and calculating the average cosine similarity distance. To embed the samples we used the \textit{multilingual-e5-base} \citep{multie5base}. As the results show, the formal tone change and typo insertion at \%10 percent result in the most semantically similar samples to the original ones. Moreover, redundant information insertion causes the samples to be most distant to the original ones. As the ambiguity and typo (\%25) inserted samples result in more performance drops then more redundant correpondents in many cases, we also show that the performances could not be entirely attributed to the semantic similarity changes.

We used the widely employed TextAttack\citep{textattack} library to generate the typos with typo-inserted perturbations. This library is frequently adopted across the literature. For example, \citep{llm_robustness2} uses TextAttack to introduce character level and word level attacks to adversarial prompts, \citep{textattackref2} uses TextAttack to introduce stochastic perturbations to text on a character level to assess the performance of text-to-image-diffusion-models-and \citep{textattackref3} uses TextAttack to generate typos to introduce a typos correction training for dense retrieval. 

\subsection{Prompts used During the Experiments}
We follow \citep{bergen} and use their prompts for question answering. From their benchmark, we used the following prompt for the experiments performed in a closed-book setting without any document insertion:

\begin{quote}
\textbf{system\_without\_docs:} "You are a helpful assistant. Answer the questions as briefly as possible."
\textbf{user\_without\_docs:} f"Question:\textbackslash\ \{question \}"
\end{quote}

For the RAG experiments where the LLMs are expected to generate an answer using the knowledge contained in the retrieved documents, we used the following prompt:

\begin{quote}
\textbf{system}: "You are a helpful assistant. Your task is to extract relevant information from provided documents and to answer to questions as briefly as possible."\\
\textbf{user}: f"Background:\symbol{92}n\{docs\}\symbol{92}n\symbol{92}nQuestion:\symbol{92} \{question\}"
\end{quote}

For the BGE BASE model, we used the following prompt given in Pyserini regressions to encode the passages:

\begin{center}
"Represent this sentence for searching relevant passages:"
\end{center}

\subsection{Details of the Answer Label Matching}
BEIR benchmark does not originally incorporate the labels of the Question Answering to their evaluation process. In order to use these dataset for the RAG setting, we collected the respective answer labels of the queries from various resources.\\
For HotpotQA we collected the answer labels from the metadata information stored within the sample instances. Similarly for the BioASQ, we followed the instructions provided by the BEIR benchmark to form the  corpus and the test set. Out of 500 test queries provided, the ones belonging to the category "Summary" are eliminated as these provide a free-form string as the reference answer. Remaining 378 questions are used and the "exact" asnwers provided are used as the golden answer of the system during the experiments. For the NQ dataset, the version contained within the BEIR benchmark is collected from the development set of the original Natural Questions \citep{nqdataset} set. In order to match the labels to the queries, we collected the subset of samples in the NQ that has a corresponding answer label in the development set.

\subsection{Retriever Robustness}

\begin{table*}[!ht]
\begin{tabular}{|l|l|l|l|l|l|l|l|}
\hline
\multicolumn{1}{|c|}{\textbf{Topic}} & \multicolumn{1}{c|}{\textbf{Retriever}} & \multicolumn{1}{c|}{\textbf{Original}} & \multicolumn{1}{c|}{\textbf{Redundancy}} & \multicolumn{1}{c|}{\textbf{Formal Tone}} & \multicolumn{1}{c|}{\textbf{Ambiguity}} & \multicolumn{1}{c|}{\textbf{T\%10}} & \multicolumn{1}{c|}{\textbf{T\%25}} \\ \hline
HotpotQA&BGE Base&71.82$\uparrow$ &66.92&69.34&64.45&62.94&47.75$\downarrow$\\ \hline
HotpotQA &Contriever&60.84$\uparrow$&59.12&59.34&56.42&53.37&39.06$\downarrow$\\ \hline
HotpotQA &BM25 Flat&60.81$\uparrow$&44.72&54.20&49.38&54.34&41.92$\downarrow$\\ \hline
HotpotQA &BM25 MF&58.0$\uparrow$&47.20&53.90&50.17&50.59&37.36$\downarrow$\\ \hline
NQ &BGE Base&64.59$\uparrow$&55.10&61.65&51.60&50.04&34.35$\downarrow$ \\ \hline
NQ &Contriever&58.60$\uparrow$&52.11&56.58&47.33&45.39&30.95$\downarrow$ \\ \hline
NQ &BM25 Flat&35.87$\uparrow$&23.64&32.80&23.55&25.87&17.12$\downarrow$ \\ \hline
NQ &BM25 MF&38.91$\uparrow$&30.38&37.68&28.57&29.35&19.73$\downarrow$ \\ \hline
BioASQ &BGE Base&36.06$\uparrow$&33.01&34.82&30.24&30.43&27.83$\downarrow$ \\ \hline
BioASQ &Contriever&34.93$\uparrow$&30.57&31.60&28.87&28.12&25.16$\downarrow$ \\ \hline
BioASQ &BM25 Flat&45.22$\uparrow$&25.01$\downarrow$&37.89&33.37&35.94&29.87 \\ \hline
BioASQ &BM25 MF&39.33$\uparrow$&25.34$\downarrow$&34.54&30.31&32.40&29.04 \\ \hline
\end{tabular}
\caption{\label{tab: remaining_retriever_results}The average retriever performances reported with metric Recall@5 (\%). The up and down arrows define the maximum and minimum performing cases, respectively. (T\%X: Typo insertion at \%X level)}
\end{table*}

\begin{figure*}[p]
\centering
\centerline{\includegraphics[width=1\textwidth]{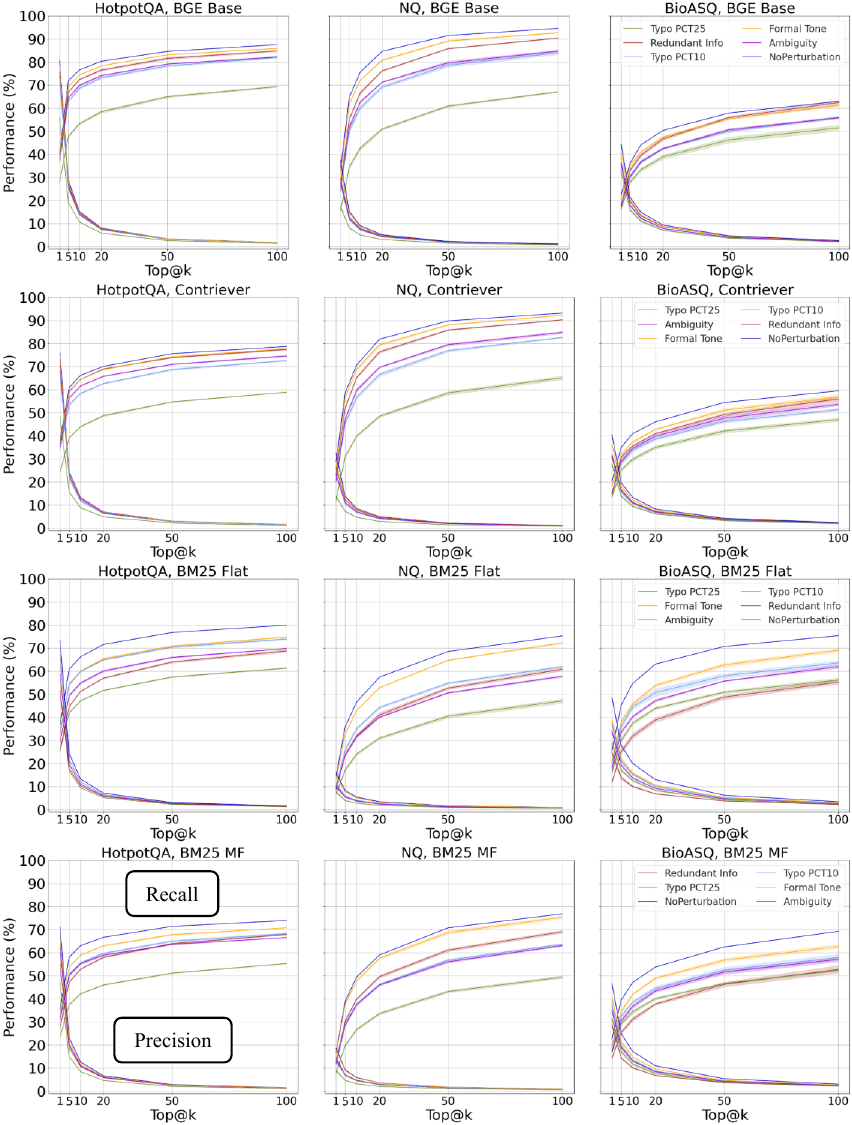}}
\caption{Remaining retriever performances with Recall@k and Precision@5 metrics on all datasets with respect to changing Top@k values.}
\label{fig: remaining_retriever_results}
\end{figure*}

For the remaining dataset and retriever combinations, the average retriever performances with different Topk@k values can be seen on Figure \ref{fig: remaining_retriever_results} while the average retriever results displayed on Figure \ref{rag_results_hotpotqa} and \ref{all_rag_match} are reported in Table \ref{tab: remaining_retriever_results}.

\subsection{RAG Robustness}

In this section of the Appendix we report the results of the experiments and results of the analysis we perform to understand the importance of parameter "k" selection and the impact of the different perturbations on different Top@K levels.

\begin{figure}[h]
\centering
\centerline{\includegraphics[width=0.5\textwidth]{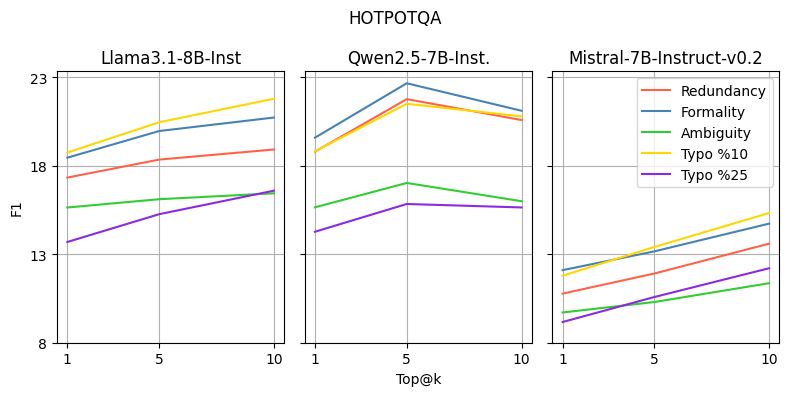}}
\caption{Effect of Top@k choice on RAG performance under different perturbations.}
\label{topkeffect}
\end{figure}

\textbf{Top-k Effect:} To understand the effect of the perturbations better, we also investigated their relationship increasing k parameter. Increasing the number of documents, i.e. k, increases the likelihood of retrieving the related text passage and returning it within the retrieved set "Top-k", however, the increase in this parameter also increases the proportion of the irrelevant documents that are returned. This is due to limited number of existing relevant documents defined per query in the system. When combined with different perturbation types, each perturbation results in different trends as shown in Figure \ref{topkeffect} with the unigram token overlap F1 used as the metric.

\begin{figure*}[h]
\centering
\centerline{\includegraphics[width=1\textwidth]{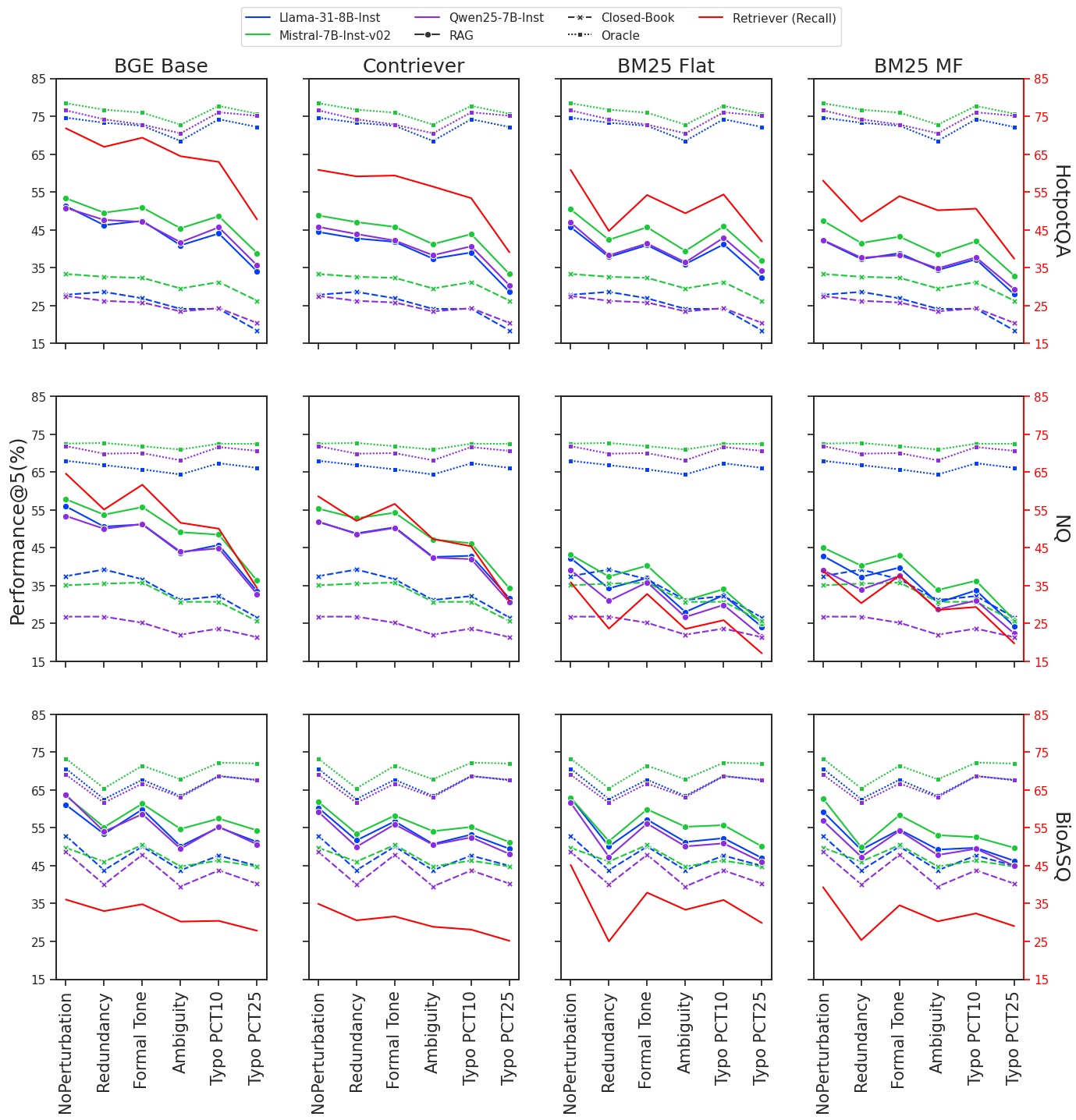}}
\caption{End-to-end RAG results with all combinations using the "Match" metric.}
\label{all_rag_match}
\end{figure*}

The end-to-end RAG performances for all generators and perturbation combinations on HotpotQA dataset, where BGE Base is used as the retriever, can be seen in Figure \ref{all_rag_match} for the k values of 1, 5, and 10.

\begin{table*}[]
\begin{tabular}{|l|l|l|l|l|l|l|l|}
\hline
Generator	 &Retriever	&Type	&R	&F	&A	&T10	&T25\\ \hline
Qwen	&Contriever	 &RET	 &0.1847	 &0.2077	 &0.2693	 &0.3273	 &0.377\\ \hline
Qwen	&Contriever	 &CB	 &0.2146	 &0.06001	 &0.1749	 &0.09721	 &0.06183\\ \hline
Qwen	&Contriever	 &ORACLE	 &0.3368	 &0.1738	 &0.3184	 &0.04723	 &0.05196\\ \hline
Mistral	&Contriever	 &RET	 &0.1854	 &0.2068	 &0.2823	 &0.2471	 &0.3502\\ \hline
Mistral	&Contriever	 &CB	 &0.2028	 &0.08272	 &0.156	 &0.07278	 &0.1357\\ \hline
Mistral	&Contriever	 &ORACLE	 &0.2558	 &0.09751	 &0.2282	 &0.0117	 &0.06488\\ \hline
Llama	&Contriever	 &RET	 &0.1659	 &0.1976	 &0.2668	 &0.2525	 &0.331\\ \hline
Llama	&Contriever	 &CB	 &0.1619	 &0.06978	 &0.172	 &0.0741	 &0.09214\\ \hline
Llama	&Contriever	 &ORACLE	 &0.2578	 &0.08921	 &0.2952	 &0.03678	 &0.05382\\ \hline
Qwen	&BM25MF	 &RET	 &0.2322	 &0.2098	 &0.2387	 &0.27	 &0.2814\\ \hline
Qwen	&BM25MF	 &CB	 &0.2241	 &0.07222	 &0.1818	 &0.157	 &0.153\\ \hline
Qwen	&BM25MF	 &ORACLE	 &0.2747	 &0.1552	 &0.3591	 &0.05237	 &0.04454\\ \hline
Mistral	&BM25MF	 &RET	 &0.2636	 &0.2129	 &0.2914	 &0.2391	 &0.2965\\ \hline
Mistral	&BM25MF	 &CB	 &0.1805	 &0.01016	 &0.1168	 &0.08652	 &0.1226\\ \hline
Mistral	&BM25MF	 &ORACLE	 &0.2685	 &0.1419	 &0.2202	 &0.1069	 &0.04519\\ \hline
Llama	&BM25MF	 &RET	 &0.2352	 &0.1939	 &0.2424	 &0.272	 &0.2544\\ \hline
Llama	&BM25MF	 &CB	 &0.09964	 &0.06859	 &0.2086	 &0.1637	 &0.08943\\ \hline
Llama	&BM25MF	 &ORACLE	 &0.2086	 &0.1104	 &0.2522	 &0.07165	 &0.1068\\ \hline
Qwen	&BM25Flat	 &RET	 &0.27	 &0.1578	 &0.2374	 &0.3436	 &0.3388\\ \hline
Qwen	&BM25Flat	 &CB	 &0.2514	 &0.08932	 &0.1691	 &0.1846	 &0.1847\\ \hline
Qwen	&BM25Flat	 &ORACLE	 &0.3248	 &0.08844	 &0.3151	 &0.03433	 &0.08664\\ \hline
Mistral	&BM25Flat	 &RET	 &0.2564	 &0.215	 &0.2625	 &0.2874	 &0.3723\\ \hline
Mistral	&BM25Flat	 &CB	 &0.2025	 &0.01857	 &0.112	 &0.1005	 &0.1469\\ \hline
Mistral	&BM25Flat	 &ORACLE	 &0.3166	 &0.1379	 &0.2291	 &0.1664	 &0.04382\\ \hline
Llama	&BM25Flat	 &RET	 &0.2732	 &0.209	 &0.3008	 &0.3086	 &0.3499\\ \hline
Llama	&BM25Flat	 &CB	 &0.108	 &0.09247	 &0.2003	 &0.1537	 &0.1534\\ \hline
Llama	&BM25Flat	 &ORACLE	 &0.1752	 &0.1043	 &0.2583	 &0.07364	 &0.1052\\ \hline
Qwen	&BGE BASE	 &RET	 &0.1041	 &0.1064	 &0.2044	 &0.2563	 &0.2753\\ \hline
Qwen	&BGE BASE	 &CB	 &0.2465	 &0.04407	 &0.152	 &0.07833	 &0.1161\\ \hline
Qwen	&BGE BASE	 &ORACLE	 &0.3899	 &0.196	 &0.323	 &0.002851  &0.1028\\ \hline
Mistral	&BGE BASE	 &RET	 &0.1145	 &0.1364	 &0.1923	 &0.248	 &0.2611\\ \hline
Mistral	&BGE BASE	 &CB	 &0.2446	 &0.09247	 &0.09016	 &0.07595	 &0.07129\\ \hline
Mistral	&BGE BASE	 &ORACLE	 &0.3205	 &0.1771	 &0.255	 &0.02001	 &0.04105\\ \hline
Llama	&BGE BASE	 &RET	 &0.05491	 &0.03533	 &0.148	 &0.2053	 &0.2333\\ \hline
Llama	&BGE BASE	 &CB	 &0.2061	 &0.08109	 &0.2326	 &0.04628	 &0.09622\\ \hline
Llama	&BGE BASE	 &ORACLE	 &0.3504	 &0.1534	 &0.3347	 &0.0413	 &0.1152\\ \hline
\end{tabular}
\caption{Pearson correlation coefficients calculated for the BioASQ dataset.}
\label{pearsoncorr_all_bioasq}
\end{table*}

\begin{table*}[]
\begin{tabular}{|l|l|l|l|l|l|l|l|}
\hline
Generator	&Retriever	&Type	&R	&F	&A	&T10	&T25\\ \hline
Qwen	&Contriever	&RET	&0.3509	&0.3041	&0.3503	&0.3869	&0.4345\\ \hline
Qwen	&Contriever	&CB	&0.02354	&0.05745	&0.1054	&0.03753	&0.08735\\ \hline
Qwen	&Contriever	&ORACLE	&0.1494	&0.1019	&0.1117	&0.05009	&0.02609\\ \hline
Mistral	&Contriever	&RET	&0.3159	&0.2952	&0.3616	&0.3825	&0.4073\\ \hline
Mistral	&Contriever	&CB	&0.0609	&0.03769	&0.1422	&0.0904	&0.1606\\ \hline
Mistral	&Contriever	&ORACLE	&0.06711	&0.09729	&0.069	&0.02333	&0.03128\\ \hline
Llama	&Contriever	&RET	&0.3173	&0.2756	&0.3344	&0.3705	&0.408\\ \hline
Llama	&Contriever	&CB	&0.02399	&0.0195	&0.1086	&0.08741	&0.1575\\ \hline
Llama	&Contriever	&ORACLE	&0.1481	&0.138	&0.1456	&0.04362	&0.05309\\ \hline
Qwen	&BM25MF	&RET	&0.4043	&0.3523	&0.4038	&0.423	&0.4728\\ \hline
Qwen	&BM25MF	&CB	&0.02542	&0.03558	&0.08528	&0.0515	&0.07004\\ \hline
Qwen	&BM25MF	&ORACLE	&0.0776	&0.08356	&0.1216	&0.02084	&0.03255\\ \hline
Mistral	&BM25MF	&RET	&0.4157	&0.3402	&0.4112	&0.4076	&0.4634\\ \hline
Mistral	&BM25MF	&CB	&0.02981	&0.04956	&0.1025	&0.07451	&0.1527\\ \hline
Mistral	&BM25MF	&ORACLE	&0.06964	&0.05083	&0.0801	&0.02289	&0.02811\\ \hline
Llama	&BM25MF	&RET	&0.3719	&0.3102	&0.3983	&0.3869	&0.4448\\ \hline
Llama	&BM25MF	&CB	&0.057	&0.06124	&0.1495	&0.09783	&0.1876\\ \hline
Llama	&BM25MF	&ORACLE	&0.1062	&0.06355	&0.087	&0.01338	&0.02591\\ \hline
Qwen	&BM25Flat	&RET	&0.4692	&0.4339	&0.4457	&0.4339	&0.4387\\ \hline
Qwen	&BM25Flat	&CB	&0.008238	&0.01181	&0.08123	&0.05578	&0.05378\\ \hline
Qwen	&BM25Flat	&ORACLE	&0.04644	&0.06749	&0.06901	&0.005686	&0.02614\\ \hline
Mistral	&BM25Flat	&RET	&0.4206	&0.3799	&0.4176	&0.3853	&0.3979\\ \hline
Mistral	&BM25Flat	&CB	&0.03037	&0.03684	&0.09031	&0.0329	&0.1377\\ \hline
Mistral	&BM25Flat	&ORACLE	&0.08621	&0.06272	&0.06035	&0.01919	&0.03969\\ \hline
Llama	&BM25Flat	&RET	&0.4136	&0.3657	&0.3794	&0.3943	&0.3911\\ \hline
Llama	&BM25Flat	&CB	-&0.0004271	&0.01917	&0.08938	&0.07803	&0.1281\\ \hline
Llama	&BM25Flat	&ORACLE	&0.06052	&0.07783	&0.09723	-&0.006478	&0.0324\\ \hline
Qwen	&BGE BASE	&RET	&0.3007	&0.2689	&0.3187	&0.3404	&0.394\\ \hline
Qwen	&BGE BASE	&CB	&0.04057	&0.05453	&0.1134	&0.04707	&0.1205\\ \hline
Qwen	&BGE BASE	&ORACLE	&0.1096	&0.1084	&0.1297	&0.0218	&0.05522\\ \hline
Mistral	&BGE BASE	&RET	&0.3047	&0.2659	&0.3275	&0.3516	&0.4085\\ \hline
Mistral	&BGE BASE	&CB	&0.07274	&0.05995	&0.1364	&0.08363	&0.1611\\ \hline
Mistral	&BGE BASE	&ORACLE	&0.05492	&0.06381	&0.08311	&0.02705	&0.01763\\ \hline
Llama	&BGE BASE	&RET	&0.3062	&0.2745	&0.3007	&0.348	&0.3977\\ \hline
Llama	&BGE BASE	&CB	&0.03287	&0.04093	&0.1064	&0.07626	&0.1631\\ \hline
Llama	&BGE BASE	&ORACLE	&0.1148	&0.143	&0.151	&0.05786	&0.03431\\ \hline\hline
\end{tabular}
\caption{Pearson correlation coefficients calculated for the NQ dataset.}
\label{pearsoncorr_all_nq}
\end{table*}

\end{document}